\documentclass[12pt,useAMS,usenatbib,nomarkers,referee]{biom}

\usepackage[T1]{fontenc}
\usepackage[utf8]{inputenc}
\usepackage{lmodern}
\usepackage{textcomp}
 
\usepackage{adjustbox}

\usepackage{fancyhdr}
\usepackage{natbib}
\usepackage{amsmath,amssymb,bm,dsfont,mathrsfs}
\usepackage[mathscr]{euscript}
\allowdisplaybreaks   
\usepackage{thmtools,thm-restate}
\usepackage{bm}
\usepackage{graphicx}
\graphicspath{{./figs/}}
 
\usepackage{float}
\usepackage{subcaption}   
\usepackage{amsfonts}
\usepackage{algorithm}
\usepackage{setspace}
\usepackage{enumitem}
\usepackage{titlesec}
\titlespacing*{\section}{0pt}{.25\baselineskip}{.1\baselineskip}
\titlespacing*{\subsection}{0pt}{.25\baselineskip}{.1\baselineskip}
\titlespacing*{\subsubsection}{0pt}{0.25em}{0.25em}

\usepackage{booktabs}
\usepackage{multirow}
\usepackage{tabularx}
\usepackage{threeparttable}
\usepackage{pdflscape}
 
\usepackage[table]{xcolor}
\makeatletter
\@ifundefined{thead}{}{}
\makeatother
\usepackage{makecell}

\usepackage{siunitx}
\newcolumntype{L}{>{\raggedright\arraybackslash}X}
\newcolumntype{C}{>{\centering\arraybackslash}m{0.9cm}} 
\newcolumntype{A}{S[table-format=1.3]} 
\newcolumntype{O}{S[table-format=1.3]} 
\newcolumntype{R}{S[table-format=1.2]} 

\usepackage{authblk}
\usepackage{multicol}
\usepackage{comment}
\usepackage[normalem]{ulem} 
\usepackage{url}
\usepackage{tikz}
\usetikzlibrary{arrows.meta, positioning, shapes.geometric}

\usepackage[hidelinks]{hyperref}

\hypersetup{
    colorlinks=true,
    urlcolor=blue,
    linkcolor=blue,
    citecolor=blue
}

\usepackage{authblk}
\usepackage{orcidlink}
\usepackage{algorithm}
\usepackage{algpseudocode} 

\algrenewcommand\algorithmicrequire{\textbf{Input:}}
\algrenewcommand\algorithmicensure{\textbf{Output:}}

\title{Adaptive Multi-Scale Forecasting and Gate-Localized Conformal Prediction for Multivariate Nonstationary Time Series}

\author{
  Ziling Ma\orcidlink{0009-0006-0323-6467}\textsuperscript{1,2,*}\thanks{Correspondence: ziling.ma@kaust.edu.sa}, %
  Junshu Jiang\orcidlink{0009-0006-5249-8949}\textsuperscript{1}, %
  Ángel López-Oriona\orcidlink{0000-0003-1456-7342}\textsuperscript{1}, %
  Ying Sun\orcidlink{0000-0001-6703-4270}\textsuperscript{1}, %
  Hernando Ombao\orcidlink{0000-0001-7020-8091}\textsuperscript{1,2}%

\textsuperscript{1}
Statistics Department, Computer, Electrical and Mathematical Sciences and Engineering Division,
King Abdullah University of Science and Technology, Thuwal 23955-6900, Saudi Arabia

\textsuperscript{2}
NeuroAI Laboratory, Computer, Electrical and Mathematical Sciences and Engineering Division,
King Abdullah University of Science and Technology, Thuwal 23955-6900, Saudi Arabia

}

\date{}

\begin{document}
\pagerange{\pageref{firstpage}--\pageref{lastpage}} \pubyear{2024}

\label{firstpage}


\begin{abstract}
 
We propose ABF-T-GLCP, a model-agnostic framework for forecasting and uncertainty quantification in nonstationary multivariate time series. The central idea is to learn an adaptive predictive state representation for point forecasting and reuse it for conformal calibration. The forecasting module combines horizon-specific temporal experts through a learned gate and refines predictions using sparse predictive transfer across related series. The uncertainty module, Gate-Localized Conformal Prediction (GLCP), uses the learned gate state, together with temporal recency, to select locally relevant calibration residuals, thereby coupling uncertainty calibration to the predictive regimes used by the forecasting model. This shared representation allows point forecasts and prediction intervals to adapt consistently under evolving temporal dynamics while retaining the model-agnostic nature of conformal prediction and yielding approximate local coverage under mild stability conditions. Experiments on a large-scale high-frequency commodity forecasting benchmark show consistent gains in point forecasting accuracy and substantially narrower prediction intervals with empirical coverage close to the nominal level. Additional results indicate that the framework extends beyond the motivating financial application. The full code is available at \href{https://github.com/arbitraryma/moe_conformal_finance}{GitHub}.

\end{abstract}


\renewcommand{\thefootnote}{\fnsymbol{footnote}}
\maketitle
\clearpage
\pagenumbering{arabic}
\setcounter{page}{2}
\pagestyle{plain}

\section{Introduction}

Forecasting financial commodity markets is fundamental to portfolio allocation, risk management, derivative pricing, and trading. Yet reliable forecasting remains challenging because commodity markets evolve under changing macroeconomic conditions, geopolitical events, supply-chain disruptions, and investor sentiment. As market conditions evolve, temporal dependencies, cross-commodity relationships, and predictive uncertainty also change, causing forecasting models that perform well in one period to deteriorate in another. Similar challenges arise in energy systems, transportation, environmental monitoring, and industrial forecasting, where related time series exhibit heterogeneous temporal dynamics and nonstationary behavior \citep{de200625,kong2025deep,murad2025wpmixer}.
Existing forecasting methods address only part of this challenge. Multi-scale forecasting captures heterogeneous temporal dependencies but typically assumes that adaptively combining multiple experts is always beneficial \citep{wang2024timemixer,chen2024pathformer}. In practice, one forecasting structure may outperform the others under the current regime, while combining weaker experts can introduce unnecessary estimation variance. Cross-series forecasting exploits predictive relationships among related series, yet evolving dependencies make indiscriminate transfer susceptible to negative transfer. Beyond accurate point forecasting, reliable uncertainty quantification is equally important. Although conformal prediction provides distribution-free uncertainty quantification, existing methods typically define calibration relevance through temporal proximity, covariates, or external similarity rather than the forecasting model's learned predictive state \citep{gibbs2021adaptive,zaffran2022adaptive,guan2023localized,hore2025conformal}. These observations motivate adaptive forecasting and uncertainty quantification under nonstationarity.

Motivated by this perspective, we propose Adaptive Best Forecasting with Transfer and Gate-Localized Conformal Prediction (ABF-T-GLCP), a model-agnostic framework for adaptive forecasting and uncertainty quantification. Adaptive Best Forecasting (ABF) constructs multiple temporal experts and learns series-specific, time-varying scale weights through a lightweight gating network. Rather than always aggregating experts, ABF selects the validation-best expert as a stable forecasting anchor and applies multi-scale corrections only when they improve validation performance. Sparse Predictive Transfer (T) further refines forecasts by modeling residual spillovers across related series while guarding against negative transfer through validation. Finally, Gate-Localized Conformal Prediction (GLCP) localizes conformal calibration using both temporal recency and the learned forecasting-regime representation, enabling the same adaptive representation to support both forecasting and uncertainty quantification.

Although developed for financial commodity forecasting, the proposed framework is broadly applicable to forecasting problems involving related time series with heterogeneous temporal dynamics, evolving predictive relationships, and regime-dependent uncertainty. We illustrate its performance on a large-scale, high-frequency financial commodity forecasting benchmark, where it consistently improves both point forecasting accuracy and interval efficiency while maintaining empirical coverage close to the nominal level.

Our main contributions are summarized as follows:
\begin{enumerate}
\item We propose \textbf{ABF-T}, a validation-driven adaptive forecasting framework that combines adaptive multi-scale forecasting with sparse predictive transfer for nonstationary time series.

\item We introduce \textbf{Gate-Localized Conformal Prediction (GLCP)}, which localizes conformal calibration using temporal recency and learned predictive-state similarity to produce adaptive prediction intervals with approximate local coverage guarantees.

\item We demonstrate that \textbf{ABF-T-GLCP} consistently improves both forecasting accuracy and interval efficiency while maintaining empirical coverage across a large-scale high-frequency commodity forecasting benchmark.
\end{enumerate}

\section{Related Work}

\subsection{Adaptive and multi-scale forecasting.}
Recent forecasting methods learn temporal structure through patch-based
Transformers, multi-scale mixing, and mixture-of-experts architectures,
including PatchTST, TimeMixer, and Time-MoE
\citep{nie2023a,wang2024timemixer,shi2025timemoe}.
Rather than proposing a new forecasting backbone, ABF-T is a model-agnostic
framework that combines scale-specific forecasts through validation-anchored
adaptation and sparse predictive transfer. It can therefore be paired with
different statistical or neural forecasting architectures.

\subsection{Conformal prediction.}
Conformal prediction (CP) provides model-agnostic, distribution-free
uncertainty quantification with finite-sample marginal coverage under
exchangeability \citep{vovk2005algorithmic,Lei03072018}. Standard split
conformal prediction globally pools calibration residuals, which can be
inefficient when uncertainty varies across time, series, or latent regimes.
Since exact conditional coverage is generally impossible without additional
assumptions \citep{foygel2021limits}, recent work improves efficiency through
adaptive localization.

\subsection{Adaptive and time-series conformal prediction.}
Classical conformal prediction relies on exchangeability, motivating a range of
extensions for distribution shift and dependent data. Weighted CP addresses
covariate shift \citep{tibshirani2019conformal,yang2024doubly}, while localized
CP weights calibration samples by their similarity to the test point
\citep{guan2023localized}, with Random Local Conformal Prediction (RLCP)
providing a randomized extension \citep{hore2025conformal}. For time-series
forecasting, ACI, AgACI, EnbPI, sequential predictive conformal inference, and
SAOCP adapt calibration under temporal distribution shifts
\citep{gibbs2021adaptive,zaffran2022adaptive,xu2021conformal,
xu2023sequential,bhatnagar2023improved}. Recent online and multi-step
extensions further address sequential forecasting settings
\citep{wang2024online}, while conformalized quantile regression (RCQR)
provides asymmetric prediction intervals \citep{romano2019conformalized}.
These methods typically define calibration relevance through time, covariates,
or externally specified similarity. In contrast, GLCP reuses the forecasting
model's learned predictive-state representation as the localization variable,
coupling point prediction and uncertainty calibration through the same adaptive
regime representation.

\subsection{Forecast specialization and regime-aware calibration.}
Mixture-of-experts models address predictive heterogeneity through adaptive
specialization \citep{jacobs1991}, and recent forecasting methods demonstrate
the benefits of specialization in heterogeneous time-series collections
\citep{lopez2025time,qiu2025duet,ma2026forecasting}. MoE-weighted conformal
prediction (MoECP) localizes residuals using gating-vector similarity
\citep{kong2026adaptive}, but does not explicitly account for temporal
staleness. In nonstationary forecasting, relevance depends on both
\emph{which} predictive regime generated an observation and \emph{when} it
occurred. GLCP therefore combines gate similarity with temporal recency to
adapt calibration to evolving predictive regimes and distribution shifts.
\section{Methodology}
We present ABF-T-GLCP, a model-agnostic framework for forecasting and uncertainty quantification in nonstationary multivariate time series. The framework consists of Adaptive Best Forecasting (ABF), sparse Predictive Transfer (T), and Gate-Localized Conformal Prediction (GLCP). ABF and T jointly produce the final point forecasts, while GLCP provides uncertainty quantification.

\subsection{Problem Formulation}

Consider $N$ related time series
\[
\mathcal{Y}
=
\{y_{j,t}:j=1,\ldots,N,\;t=1,\ldots,T\},
\]
where $y_{j,t}\in\mathbb{R}$ denotes the observation of series $j$ at time
$t$. Let
\[
\mathbf y_t=(y_{1,t},\ldots,y_{N,t})^\top,
\qquad
\mathcal F_t=\{\mathbf y_1,\ldots,\mathbf y_t\},
\]
denote the observations across all series at time $t$ and the information
available up to time $t$, respectively. For a forecasting horizon $h$, the objective is
to predict the future value $y_{j,t+h}$ for each series
$j\in\{1,\ldots,N\}$ using only $\mathcal F_t$.

The proposed framework proceeds in two stages. First, the ABF-T forecasting model
produces a point forecast $\hat{y}_{j,t+h}$. Second, GLCP converts the point forecast into a prediction interval
\[
C_{j,t+h}
=
[L_{j,t+h},\,U_{j,t+h}],
\]
designed to satisfy the target marginal coverage
\[
\Pr\!\left(y_{j,t+h}\in C_{j,t+h}\right)
\ge 1-\alpha,
\]
where $\alpha\in(0,1)$ denotes the prescribed miscoverage level.

Unlike standard conformal prediction, which calibrates using a globally pooled
calibration set, GLCP performs localized calibration by assigning larger
weights to calibration observations that are both temporally recent and
generated under forecasting regimes similar to the current test sample. 

To prevent information leakage, the data are partitioned chronologically
according to the forecast origin (time $t$) as
\[
\textsc{Train}
\rightarrow
\textsc{Val}
\rightarrow
\textsc{Cal}
\rightarrow
\textsc{Test}.
\]
The \textsc{Train} and \textsc{Val} sets are used to fit the forecasting model and perform model selection, including anchor selection, multi-scale blending, and sparse predictive transfer. The \textsc{Cal} set is used for GLCP bandwidth selection and conformal calibration, while the \textsc{Test} set is reserved exclusively for final evaluation.

\subsection{Adaptive Best Forecasting}

To capture temporal dependencies operating at different time scales, we train a
library of forecasting experts, each associated with a different lookback
length $L_k$, where $\mathcal L=\{L_1,\ldots,L_K\}$ \citep{chen2024pathformer,wang2024timemixer}.
For forecasting horizon $h$, the expert associated with lookback window
$L_k$, where $k\in\{1,\ldots,K\}$, predicts the future observation
\[
\hat y_{t+h,j,k}
=
f_{j,k}\!\left(x_t^{(L_k)}\right),
\]
where $x_t^{(L_k)}$ denotes the scale-specific input features constructed
from the histories of all related series.

Rather than assigning fixed scale weights, we learn an asset-specific gating
network. Let $z_{t,j}\in\mathbb{R}^{d}$ denote the regime feature vector for
series $j$ at time $t$. The gate maps $z_{t,j}$ to adaptive scale weights
\begin{equation*}
    \boldsymbol{\pi}^{(h)}_{t,j}
=
g_{\theta}(z_{t,j}),
\qquad
\sum_{k=1}^{K}
\pi^{(h)}_{t,j,k}
=
1,
\end{equation*}
where $\pi^{(h)}_{t,j,k}$ denotes the weight assigned to expert $k$, $k\in\{1,\cdots,K\}$ for
forecasting series $j$ at horizon $h$.

A standard mixture-of-experts prediction is
\begin{equation*}
    \hat y^{\mathrm{gate}}_{t+h,j}
=
\sum_{k=1}^{K}
\pi^{(h)}_{t,j,k}
\hat y_{t+h,j,k}.
\end{equation*}
Although adaptive mixtures of experts are flexible, combining a strong expert
with weaker ones may introduce unnecessary estimation variance and possible bias, thereby degrading forecasting accuracy. The final ABF, $\hat y^{\mathrm{ABF}}_{t+h,j}$ selects the validation-best expert for each forecasting horizon $h$,
\begin{equation}
a^{(h)}
=
\arg\min_{k}
\mathrm{MSE}_{\mathrm{val}}
\!\left(
\hat y_{\cdot+h,\cdot,k}
\right),
\label{eq:anchor}
\end{equation}
and uses it as a stable forecasting anchor. The remaining experts contribute
only through gated deviations,

\begin{equation*}
\hat y^{\mathrm{ABF}}_{t+h,j}
=
\hat y_{t+h,j,a^{(h)}}
+
\rho^{(h)}
\sum_{k=1}^{K}
\pi^{(h)}_{t,j,k}
\left(
\hat y_{t+h,j,k}
-
\hat y_{t+h,j,a^{(h)}}
\right),
\label{eq:abf}
\end{equation*}
where the correction strength $\rho^{(h)}\in[0,1]$ is selected on the
validation set. When $\rho^{(h)}=0$, ABF reduces to the anchor forecast,
whereas $\rho^{(h)}=1$ fully applies the gate-weighted correction.
Intermediate values balance stability and adaptive multi-scale forecasting.
\subsection{Sparse predictive transfer}

Although ABF captures the dominant within-series temporal dynamics, related
series may still exhibit residual predictive associations arising from common
market factors, sector-level dependencies, or lead--lag relationships. We
therefore refine the ABF forecasts through sparse predictive transfer.

Let
\[
e^{(h)}_{t,j}
=
y_{t+h,j}
-
\hat y^{\mathrm{ABF},(h)}_{t,j}
\]
denote the ABF residual for forecasting horizon $h$. Since most temporal
variation has already been explained by ABF, the remaining cross-series
dependence is modeled through a sparse residual correction.

For each time point, we construct the transfer feature vector

\begin{equation*}
\phi^{(h)}_t
=
\mathrm{vec}
\!\left(
\{
\pi^{(h)}_{t,i,k}
\hat y^{(h)}_{t,i,k}
\}_{i=1,\ldots,N;\,k=1,\ldots,K}
\right),
\label{eq:transfer_features}
\end{equation*}

which stacks the gate-weighted expert predictions from all source series and
temporal scales. The residual correction for target series $j$ is
\[
e^{(h)}_{t,j}
\approx
b^{(h)}_{0,j}
+
(\phi^{(h)}_t)^\top
b^{(h)}_j,
\]
where $b^{(h)}_j$ is estimated using a regularized linear regression model,
such as Lasso, Ridge, or Elastic Net.
The final point forecast for horizon $h$ is
\begin{equation}
\hat y^{\mathrm{ABF\text{-}T}}_{t+h,j}
=
\hat y^{\mathrm{ABF}}_{t+h,j}
+
\gamma^{(h)}
(
b^{(h)}_{0,j}
+
(\phi^{(h)}_t)^\top
b^{(h)}_j
).
\label{eq:final_forecast}
\end{equation}
The regularization hyperparameters and the shrinkage coefficient
$\gamma^{(h)}\in[0,1]$ are selected on the validation set, where larger
$\gamma^{(h)}$ allows stronger transfer.
\subsection{Gate-localized conformal prediction}
\label{sec:glcp}

The ABF-T model provides point forecasts but not predictive uncertainty. GLCP
constructs adaptive prediction intervals by localizing conformal calibration
according to both temporal recency and the learned forecasting regime.

Let $\hat{\sigma}^{(h)}_j>0$ denote a robust local residual scale, estimated as the Median Absolute Deviation (MAD) of the recent rolling calibration residuals from the ABF-T point forecaster. The normalized conformal score is
\[
S^{(h)}_{t,j}
=
|y_{t+h,j}
-
\hat y^{\mathrm{ABF\text{-}T}}_{t+h,j}|
/
\hat{\sigma}^{(h)}_j.
\] 
Let $\mathcal H_t$ denote the historical calibration indices available
before forecast origin $t$, including the calibration set and previously
observed test samples with $g+h\le t$. Each calibration observation
$g\in\mathcal H_t$ is assigned the weight
\begin{equation}
w_g^{(h,t,j)}
=
\lambda^{\,t-g}
\exp\!\left(
-\|
\boldsymbol{\pi}^{(h)}_{g,j}
-
\boldsymbol{\pi}^{(h)}_{t,j}
\|_2^2
(2\tau^2)^{-1}
\right),
\label{eq:glcp_weight}
\end{equation}
where $\lambda\in(0,1]$ controls temporal recency and $\tau>0$ determines the
degree of gate-based localization.

The normalized weights are
\[
\bar w_g^{(h,t,j)}
=
w_g^{(h,t,j)}
/
\sum_{\ell\in\mathcal H_t}
w_\ell^{(h,t,j)},
\qquad
\sum_{g\in\mathcal H_t}
\bar w_g^{(h,t,j)}
=
1.
\]
The effective sample size (ESS),
\[
\mathrm{ESS}^{(h)}_{t,j}
=
(
\sum_{g\in\mathcal H_t}
(\bar w_g^{(h,t,j)})^2
)^{-1},
\]
where smaller ESS indicates stronger localization.

The GLCP radius $q^{(h)}_{t,j}$ is the smallest threshold whose cumulative
normalized weight reaches $1-\alpha$,
\begin{equation*}
\begin{aligned}
q^{(h)}_{t,j}
&=
Q_{1-\alpha}^{w}
\!\left(
\{S^{(h)}_{g,j}:g\in\mathcal H_t\}
\right)
\\
&=
\inf\Bigg\{
q\in\mathbb R:
\sum_{g\in\mathcal H_t}
\bar w_g^{(h,t,j)}
\mathbf 1
(S^{(h)}_{g,j}\le q)
\ge
1-\alpha
\Bigg\}.
\end{aligned}
\label{eq:weighted_quantile}
\end{equation*}

The resulting prediction interval is

\[
C^{(h),1-\alpha}_{t,j}
=
\left[
\hat y^{\mathrm{ABF\text{-}T}}_{t+h,j}
-
q^{(h)}_{t,j}\hat{\sigma}^{(h)}_j,\;
\hat y^{\mathrm{ABF\text{-}T}}_{t+h,j}
+
q^{(h)}_{t,j}\hat{\sigma}^{(h)}_j
\right].
\]

If $\mathrm{ESS}^{(h)}_{t,j}<m_{\min}$, GLCP falls back to the corresponding
unweighted conformal quantile.
\subsection{Asymmetric GLCP}

To accommodate asymmetric predictive error distributions, we calibrate the
lower and upper prediction bounds separately. Define the normalized signed
error
\begin{align*}
E^{(h)}_{t,j}
&=
\left(
y_{t+h,j}
-
\hat y^{\mathrm{ABF\text{-}T}}_{t+h,j}
\right)
/\hat{\sigma}^{(h)}_j,
\\
S^{-,(h)}_{t,j}
&=
\max(-E^{(h)}_{t,j},0),
\qquad
S^{+,(h)}_{t,j}
=
\max(E^{(h)}_{t,j},0).
\label{eq:asym_scores}
\end{align*}

Using the same localized weights as in Eq.~\eqref{eq:glcp_weight}, the lower
and upper radii are
\[
q^{-,(h)}_{t,j}
=
Q_{1-\alpha/2}^{w}
\!(
\{S^{-,(h)}_{g,j}\}
),
\:
q^{+,(h)}_{t,j}
=
Q_{1-\alpha/2}^{w}
\!(
\{S^{+,(h)}_{g,j}\}
),
\]
where the weighted empirical quantiles are computed over
$g\in\mathcal H_t$. The resulting interval is
\begin{equation*}
C^{(h),1-\alpha}_{t,j}
=
[
\hat y^{\mathrm{ABF\text{-}T}}_{t+h,j}
-
q^{-,(h)}_{t,j}\hat{\sigma}^{(h)}_j,\;
\hat y^{\mathrm{ABF\text{-}T}}_{t+h,j}
+
q^{+,(h)}_{t,j}\hat{\sigma}^{(h)}_j
].
\label{eq:asym_interval}
\end{equation*}

The complete pseudocode is provided in Algorithm~\ref{alg:method}.
\begin{algorithm}[h]
\caption{Adaptive Forecasting with Asymmetric GLCP}
\label{alg:method}
\begin{algorithmic}[1]
\Require Panel returns $\{r_{t,j}\}$, horizon $h$, scales $\mathcal L$,
miscoverage level $\alpha$
\State Construct targets $\{y_{t+h,j}\}$
\State Split:
\textsc{Train}$\rightarrow$\textsc{Val}$\rightarrow$\textsc{Cal}$\rightarrow$\textsc{Test}
\For{each scale $L_k\in\mathcal L$}
    \State Train expert $f_{j,k}$ on \textsc{Train} and obtain
    $\hat y_{t+h,j,k}$
\EndFor
\State Train gate $g_\theta(z_{t,j})$ and obtain $\boldsymbol{\pi}_{t,j}$
\State Select anchor $a$ and correction strength $\rho$ on \textsc{Val}
\State Form ABF forecasts $\hat y^{\mathrm{ABF}}_{t+h,j}$  
\State Fit transfer correction on \textsc{Val}
\State Select final forecast
$\hat y^{\mathrm{ABF\text{-}T}}_{t+h,j}$ on \textsc{Val}
using Eq.~\eqref{eq:final_forecast}
\State Tune localization bandwidth $\tau$ on \textsc{Cal}
\For{each $t\in\textsc{Test}$ and series $j$}
    \State Compute GLCP weights $w_g^{(h,t,j)}$ using Eq.~\eqref{eq:glcp_weight}
    \State Compute radii $q^{-,(h)}_{t,j}$ and $q^{+,(h)}_{t,j}$
    \State Output
    $
    [
    \hat y^{\mathrm{ABF\text{-}T}}_{t+h,j}
    -
    q^{-,(h)}_{t,j}\hat{\sigma}^{(h)}_j,\;
    \hat y^{\mathrm{ABF\text{-}T}}_{t+h,j}
    +
    q^{+,(h)}_{t,j}\hat{\sigma}^{(h)}_j
    ]
    $
    \State Update calibration history
\EndFor
\end{algorithmic}
\end{algorithm}

\subsection{Assumptions and local validity}
\label{subsec:local_validity}

Since exact distribution-free conditional coverage is impossible under
arbitrary nonstationarity without additional assumptions, we state an
approximate local validity result. The full proof is provided in the
Supplementary.

\begin{assumption}[Predictable tail-wise localization]\label{assumption}
For each forecasting horizon $h$ and test pair $(t,j)$, the weights
$w_g^{(h,t,j)}$, $g\in\mathcal H_t$, are $\mathcal F_t$-measurable and do not
depend on the unobserved response $y_{t+h,j}$. Moreover, the localized
weighted distributions of the historical lower- and upper-tail scores
approximate the corresponding test score distributions up to errors
$\delta^{-,(h)}_{t,j}$ and $\delta^{+,(h)}_{t,j}$.
\end{assumption}

Let $\widehat F_{t,j}^{-,w,(h)}$ and $\widehat F_{t,j}^{+,w,(h)}$ denote the
weighted empirical CDFs of $\{S^{-,(h)}_{g,j}:g\in\mathcal H_t\}$ and
$\{S^{+,(h)}_{g,j}:g\in\mathcal H_t\}$ under weights
$\bar w_g^{(h,t,j)}$. Suppose their deviations from the corresponding
localized population CDFs are bounded by
$\epsilon^{-,(h)}_{t,j}(\eta)$ and $\epsilon^{+,(h)}_{t,j}(\eta)$ with
probability at least $1-\eta$.

\begin{theorem}[Approximate local coverage for asymmetric GLCP]
Under Assumption~1 and the above weighted empirical concentration event, with
probability at least $1-\eta$ over the calibration history, the asymmetric
GLCP interval satisfies
\[
\begin{aligned}
\Pr
\left(
y_{t+h,j}\in C^{(h),1-\alpha}_{t,j}
\mid \mathcal F_t
\right)
&\ge
1-\alpha
-\delta^{-,(h)}_{t,j}
-\delta^{+,(h)}_{t,j}  \\
&\quad
-\epsilon^{-,(h)}_{t,j}(\eta)
-\epsilon^{+,(h)}_{t,j}(\eta).
\end{aligned}
\]
\label{the:theorem}
\end{theorem}

The bound separates localization error from finite-sample weighted estimation
error. Smaller $\delta$ terms indicate better local matching, whereas stronger
localization can reduce effective sample size and increase the $\epsilon$
terms. The symmetric GLCP guarantee follows by applying the same argument to
$S^{(h)}_{t,j}=|E^{(h)}_{t,j}|$ with a single weighted quantile at level
$1-\alpha$. As limiting cases, GLCP recovers uniform split conformal
calibration $(\lambda=1,\tau\to\infty)$, recency-weighted calibration
$(\tau\to\infty,\lambda<1)$, and gate-localized calibration
$(\lambda=1,\tau<\infty)$.
\section{Experiments} 
\subsection{Baselines.}
For point forecasting, we compare ABF-T with three multi-scale baselines:
\emph{Single-Scale}, \emph{Equal-Weight}, and \emph{Asset-Specific Gated Scale
MoE}. These baselines isolate the effects of scale selection, uniform
multi-scale averaging, and learned adaptive weighting. All methods use the
Heterogeneous Autoregressive (HAR) model as a common backbone, a standard
parsimonious benchmark for realized volatility forecasting
\citep{corsi2009simple,luo2022forecasting}.

For uncertainty quantification, we compare GLCP with representative conformal calibration baselines, all built on the same ABF-T point forecasts. Table~\ref{tab:cp_characteristics} summarizes whether each method is online, localized, MoE-based, or time-aware, where these categories indicate sequential updating, similarity-weighted calibration, use of a gating mechanism or mixture structure, and temporal recency, respectively.

\begin{table}[t]
\centering

\setlength{\tabcolsep}{5pt}
\begin{tabular}{lcccc}
\toprule
Method & Online & Localized & MoE-based & Time-aware \\
\midrule
ACI    & \checkmark &            &            & \checkmark \\
AgACI  & \checkmark &            &            & \checkmark \\
SAOCP  & \checkmark &            &            & \checkmark \\
RCQR   & \checkmark &            &            & \checkmark \\
EnbPI  & \checkmark &            &            & \checkmark \\
RLCP   & \checkmark & \checkmark &            & \checkmark \\
MoECP  &            & \checkmark & \checkmark &            \\
\midrule
GLCP
       & \checkmark & \checkmark & \checkmark & \checkmark \\
\bottomrule
\end{tabular}
\caption{Methodological characteristics of the conformal calibration methods.}
\label{tab:cp_characteristics}
\end{table}

\subsection{Evaluation metrics.}
Point forecasting performance is evaluated using Mean Squared Error (MSE) and Mean Absolute Error (MAE). For uncertainty quantification, we report empirical Coverage (Cov), average Interval Width (Width), and Interval Score (IS), which jointly evaluate calibration and sharpness. For localized conformal methods, we additionally report the Effective Sample Size (ESS). We use $m_{\min}=100$ throughout the experiments.

\subsection{Dataset and experimental setup.}
Experiments are conducted on a challenging publicly available high-frequency
Chinese commodity futures dataset spanning one year from August~1,~2022 to
August~1,~2023 \citep{jiang2025efficient}. The dataset contains
millisecond-level prices for 55 liquid commodity futures. Prices are resampled
into 5-minute bars within each trading segment, and log returns are computed
separately within each segment to avoid artificial overnight returns. Assets
with excessive missing values are removed, while the remaining missing
observations are forward- and backward-filled. This preprocessing yields
approximately $5.5\times10^4$ synchronized 5-minute observations across 55
assets. Samples are split chronologically into training (50\%), validation
(15\%), calibration (20\%), and test (15\%) sets. The gate input $z_{t,j}$ is
constructed only from information in $\mathcal F_t$, including asset-specific
multi-scale volatility summaries and scale-expert signals, with no future
outcome information. Results are averaged over 20 random seeds.

\subsection{Forecasting target.}
Following the realized volatility literature \citep{andersen2003modeling,barndorff2002econometric,liu2015does}, the forecasting target is the future log realized volatility. Let
\[
r_{j,t}=\log P_{j,t}-\log P_{j,t-1}
\]
denote the 5-minute log return of asset $j$. For forecasting horizon $h$, the target is
\[
y_{j,t+h}
=
\log (
\sum_{\ell=1}^{h} r_{j,t+\ell}^{2}
+\epsilon
 ),
\]
where $\epsilon>0$ is a small constant. We evaluate forecasting horizons $h\in\{3,12,24,48,96\}$, corresponding to approximately 15 minutes, 1 hour, 2 hours, 4 hours, and 8 hours of observed trading time. Horizons are defined over observed trading bars and may span market breaks, while returns are always computed within trading sessions to avoid artificial jumps. We consider the lookback windows $L\in\{1,3,6,12,24,48,96,240\}$.

\subsection{Analysis of point forecast results }
Table~\ref{tab:point_best} compares ABF-T with the strongest baseline at each forecasting horizon. ABF-T consistently achieves the lowest MSE and MAE across all horizons, with the improvements becoming more pronounced as the forecasting horizon increases. This suggests that the proposed anchor-based blending strategy is particularly effective as forecasting uncertainty grows and adaptive multi-scale information becomes increasingly valuable. By anchoring predictions to the validation-best expert while introducing only gated, regime-dependent corrections, ABF-T reduces unnecessary estimation variance without sacrificing adaptability, yielding more robust long-horizon forecasts.

\begin{table}[!t]
\centering

\setlength{\tabcolsep}{4pt}
\begin{tabular}{c cc cc cc}
\toprule
\multirow{2}{*}{$h$}
& \multicolumn{2}{c}{Best Baseline}
& \multicolumn{2}{c}{ABF-T}
& \multicolumn{2}{c}{Gain (\%)} \\
\cmidrule(lr){2-3}
\cmidrule(lr){4-5}
\cmidrule(lr){6-7}
& MSE & MAE & MSE & MAE & MSE$\downarrow$ & MAE$\downarrow$ \\
\midrule
3  & 0.862 & 0.661 & \textbf{0.851} & \textbf{0.655} & \textbf{1.27} & \textbf{1.00} \\
12 & 0.628 & 0.599 & \textbf{0.601} & \textbf{0.584} & \textbf{4.38} & \textbf{2.47} \\
24 & 0.883 & 0.721 & \textbf{0.828} & \textbf{0.687} & \textbf{6.29} & \textbf{4.66} \\
48 & 1.946 & 1.105 & \textbf{1.752} & \textbf{1.043} & \textbf{10.00} & \textbf{5.59} \\
96 & 3.848 & 1.528 & \textbf{3.270} & \textbf{1.410} & \textbf{15.02} & \textbf{7.71} \\
\bottomrule
\end{tabular}
\caption{Point forecasting performance of ABF-T compared with the strongest non-ABF baseline at each forecasting horizon. Relative improvements are computed with respect to the best baseline selected by MSE. Lower values are better.}
\label{tab:point_best}
\end{table}

\subsection{Analysis of conformal prediction results }
\begin{figure*}[t]
    \centering
    \includegraphics[width=1\linewidth]{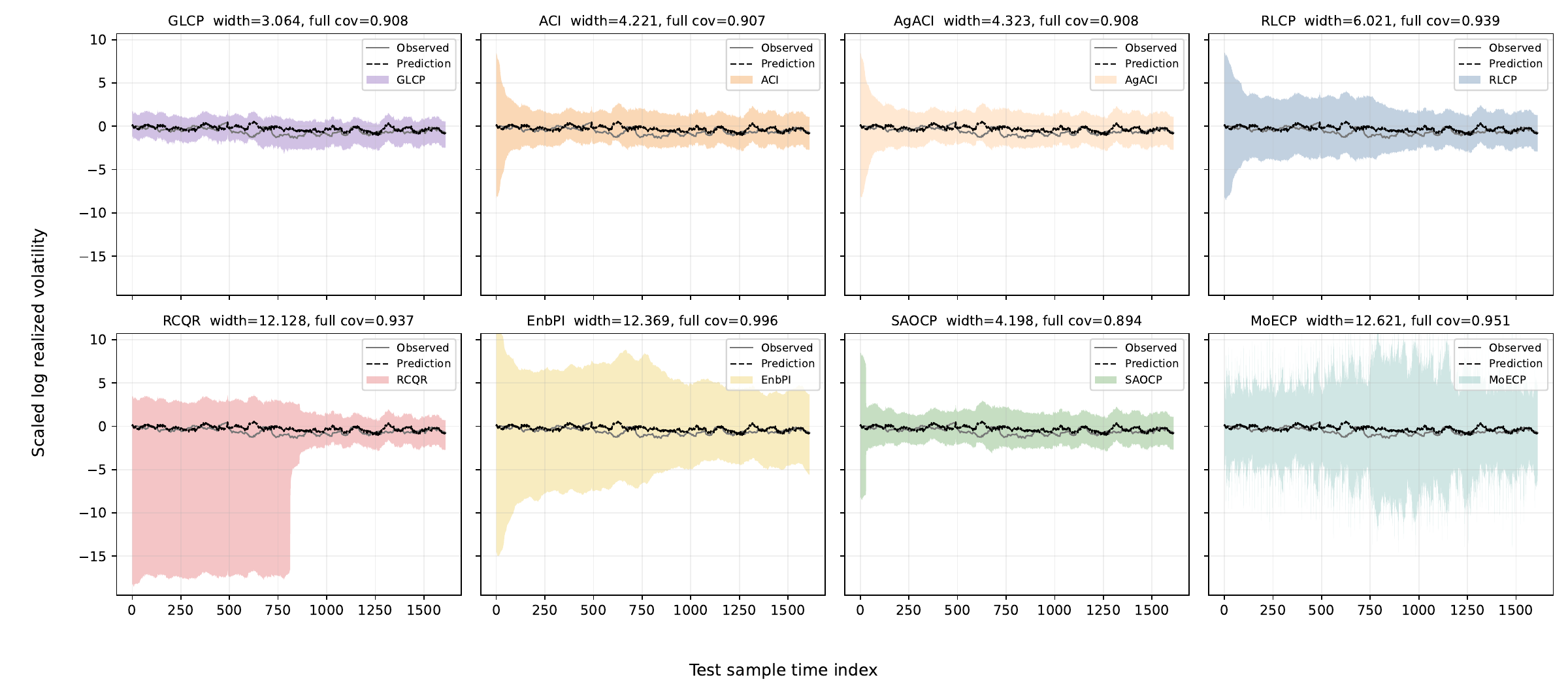}
    \caption{Prediction intervals at horizon $h=48$. GLCP maintains near-nominal coverage with visibly narrower intervals than competing conformal baselines.}
\label{fig:interval_panels_h48}
\end{figure*}

Table~\ref{tab:uq_results} compares the calibration--efficiency trade-offs of the conformal prediction methods. GLCP consistently achieves the lowest interval scores and the narrowest prediction intervals while maintaining empirical coverage close to the nominal 90\% level, with only slight undercoverage at $h=12$ and $h=24$. Its average ESS lies between global calibration (ESS $=2000$) and highly localized calibration (RLCP, ESS $\approx3$), indicating effective localization without sacrificing substantial calibration information. Consistent with the average ESS of 304, the minimum ESS threshold ($m_{\min}=100$) was never reached, and the fallback to unweighted conformal calibration was therefore never triggered. Figure~\ref{fig:interval_panels_h48} illustrates the prediction intervals for a representative test segment at horizon $h=48$, showing the sharper intervals produced by GLCP. The dotted line denotes the ABF-T point forecast.

\begin{table*}[t]
\centering
\footnotesize
\setlength{\tabcolsep}{2.5pt}
\begin{tabular}{l ccc ccc ccc ccc ccc c}
\toprule
\multirow{2}{*}{Method}
& \multicolumn{3}{c}{$h=3$}
& \multicolumn{3}{c}{$h=12$}
& \multicolumn{3}{c}{$h=24$}
& \multicolumn{3}{c}{$h=48$}
& \multicolumn{3}{c}{$h=96$}
& \multirow{2}{*}{ESS} \\
\cmidrule(lr){2-4}
\cmidrule(lr){5-7}
\cmidrule(lr){8-10}
\cmidrule(lr){11-13}
\cmidrule(lr){14-16}
& Cov & Width & IS
& Cov & Width & IS
& Cov & Width & IS
& Cov & Width & IS
& Cov & Width & IS
& \\
\midrule

ACI
& 0.900 & 2.75 & 3.98
& 0.901 & 2.47 & 3.08
& 0.902 & 2.71 & 3.35
& 0.907 & 4.22 & 4.79
& 0.914 & 5.32 & 5.85
& 2000 \\

AgACI
& 0.900 & 2.76 & 3.99
& 0.900 & 2.48 & 3.11
& 0.902 & 2.73 & 3.39
& 0.908 & 4.32 & 4.92
& 0.916 & 5.50 & 6.04
& 2000 \\

RLCP
& 0.921 & 3.11 & 4.18
& 0.933 & 3.10 & 3.60
& 0.929 & 3.40 & 3.96
& 0.939 & 6.02 & 6.57
& 0.946 & 8.85 & 9.51
& 3.4 \\

RCQR
& 0.916 & 3.03 & 4.09
& 0.922 & 3.95 & 4.45
& 0.925 & 4.12 & 4.64
& 0.937 & 12.13 & 12.61
& 0.944 & 10.65 & 11.29
& 2000 \\

EnbPI
& 0.975 & 4.99 & 5.48
& 0.992 & 6.08 & 6.22
& 0.997 & 7.72 & 7.82
& 0.996 & 12.37 & 12.41
& 0.989 & 12.36 & 12.50
& 2000 \\

SAOCP
& 0.894 & 2.70 & 4.01
& 0.893 & 2.47 & 3.18
& 0.891 & 2.73 & 3.49
& 0.894 & 4.20 & 4.99
& 0.897 & 5.35 & 6.28
& 896 \\

MoECP
& 0.922 & 3.25 & 4.32
& 0.917 & 4.11 & 4.72
& 0.925 & 5.48 & 6.08
& 0.951 & 12.62 & 13.10
& 0.959 & 12.61 & 13.20
& 715 \\

\midrule

GLCP (Sym.)
& 0.898 & 2.71 & 3.94
& 0.895 & 2.35 & 2.99
& 0.896 & 2.62 & 3.28
& 0.913 & 4.17 & 4.66
& 0.908 & 5.87 & 6.42
& \multirow{2}{*}{304} \\

\textbf{GLCP (Asym.)}
& {0.899} & {2.59} & \textbf{3.72}
& 0.893 & {2.12} & \textbf{2.69}
& 0.895 & {2.14} & \textbf{2.70}
& 0.908 & {3.06} & \textbf{3.48}
& 0.902 & {3.49} & \textbf{3.94}
&   \\

\midrule

Gain vs. best
& -- & -- & 6.5\%
& -- & -- & 12.8\%
& -- & -- & 19.6\%
& -- & -- & 27.3\%
& -- & --  & 32.6\%
& -- \\

\bottomrule
\end{tabular}
\caption{Uncertainty quantification results across forecasting horizons. ``GLCP (Sym.)'' and ``GLCP (Asym.)'' denote the symmetric and asymmetric variants of the proposed GLCP framework, respectively. Lower IS indicate better uncertainty quantification.}
\label{tab:uq_results}
\end{table*}

\subsection{Asset-specific expert gate-weight dynamics}

A central motivation for the proposed forecasting component is that related assets need not share the same predictive time scale. In commodity markets, short-memory dynamics may be informative for some assets during rapidly changing regimes, whereas longer-memory volatility patterns may be more useful for others. A static averaging rule over experts at different scales can therefore obscure cross-asset differences and dilute strong asset-specific predictors.

\begin{figure}[!t]
    \centering
    \includegraphics[width=\textwidth]{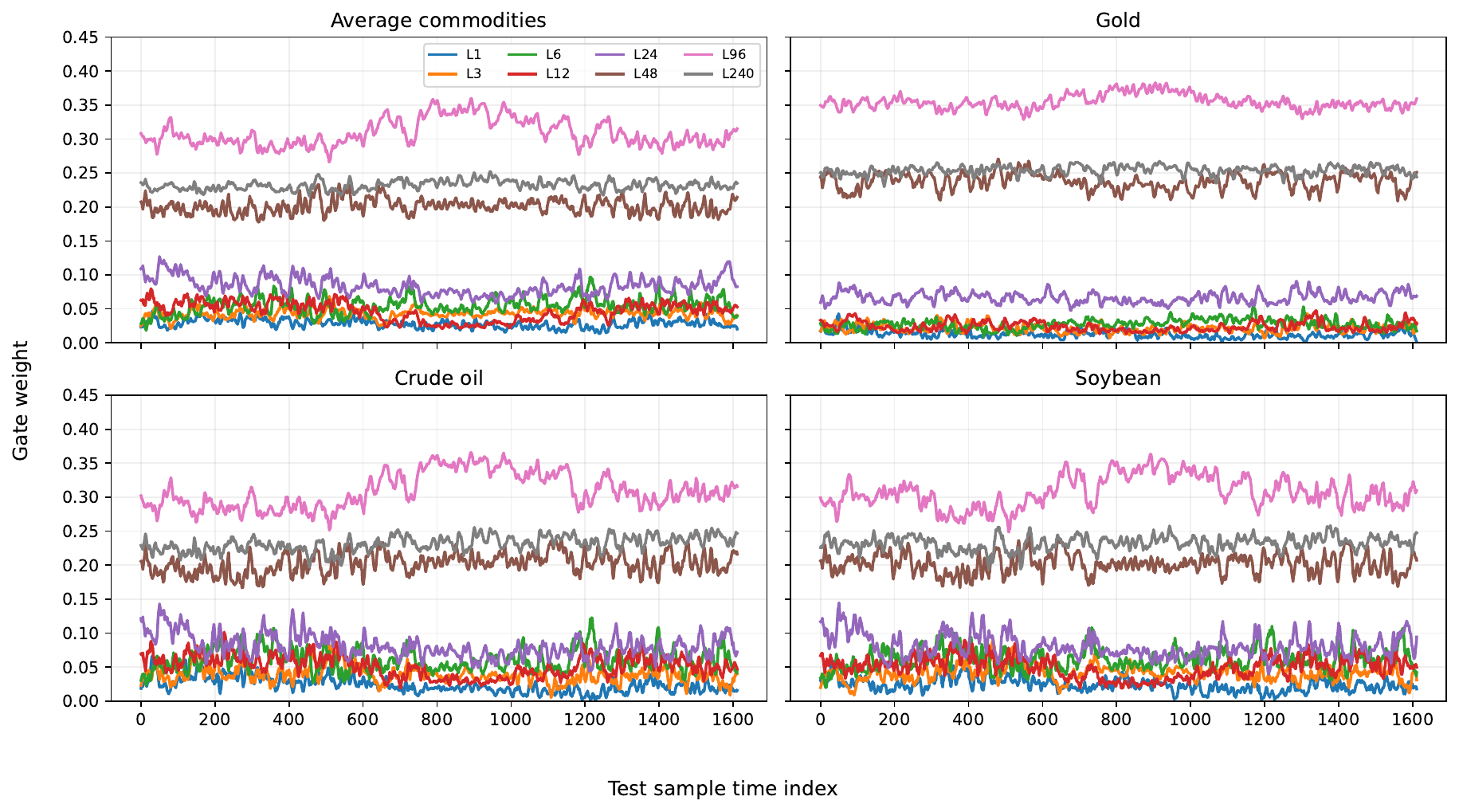}
    \caption{Gate-weight dynamics for experts at horizon $h=48$. The panels compare the average gate across all commodities with commodity-specific gates for gold, crude oil, and soybean.}
    \label{fig:gate_asset_specific}
\end{figure}

\begin{figure}[!t]
    \centering
    \includegraphics[width=\textwidth]{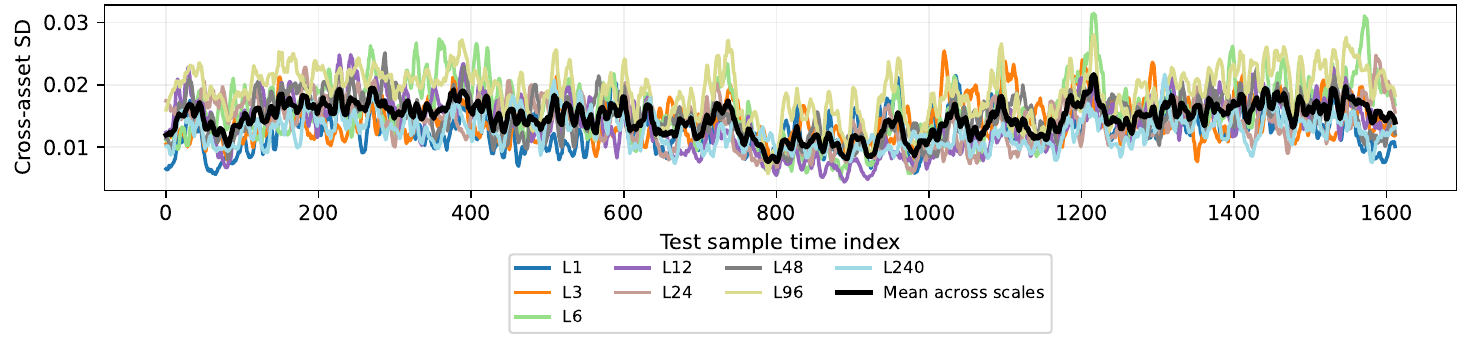}
    \caption{Cross-asset dispersion of gate weights for experts. Persistent dispersion indicates that commodities rely on different predictive time scales.}
    \label{fig:gate_cross_asset_sd}
\end{figure}

Figure~\ref{fig:gate_asset_specific} illustrates the learned gate dynamics for $h=48$. The average gate across all commodities exhibits a stable hierarchy of temporal scales, while representative commodities (gold, crude oil, and soybean) display distinct time-varying scale preferences. This suggests that the proposed asset-specific gating network captures both shared market regimes and commodity-specific temporal dynamics. Figure~\ref{fig:gate_cross_asset_sd} further confirms persistent cross-asset heterogeneity in the learned gate weights.

\subsection{Ablation studies}
Additional ablations, sensitivity analyses, full baseline results, standard deviations, hyperparameter selection, and implementation details are provided in the Supplementary.
 
\begin{figure}[!t]
    \centering
    \includegraphics[width=1\linewidth]{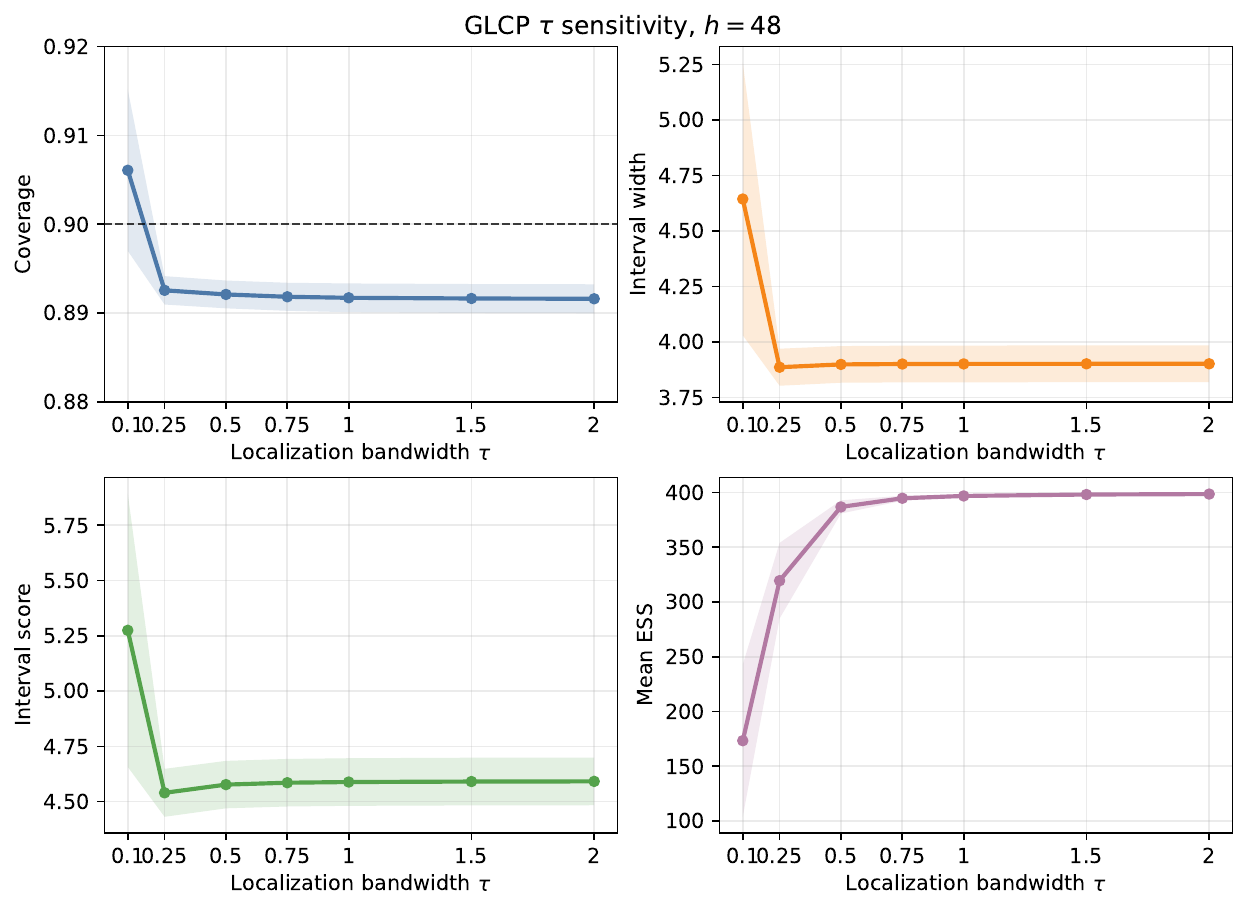}
    \caption{Influence of the selection of $\tau$ on the test set when $h=48$.}
    \label{fig:tau_sen}
\end{figure}

\paragraph{Sensitivity analysis: $\tau$.}
The localization bandwidth $\tau$ is selected using the \textsc{Cal} set only. Figure~\ref{fig:tau_sen} shows GLCP performance on the \textsc{Test} set over a range of fixed $\tau$ values for the representative horizon $h=48$, solely to illustrate sensitivity. The \textsc{Test} set is never used for bandwidth selection.

\paragraph{Neural backbone and component ablations.}
Table~\ref{tab:patchtst_modularity_glcp} uses PatchTST as a strong neural backbone to test whether ABF-T-GLCP remains effective beyond HAR experts. The point-forecast rows isolate the effects of adaptive aggregation and cross-series transfer: ABF-PatchTST adds adaptive multi-scale aggregation, while ABF-T-PatchTST further adds the transfer layer and attains the lowest MSE across the reported horizons. The UQ rows isolate the contributions of temporal recency, gate similarity, and full GLCP. The full variant achieves the lowest interval score, indicating that recency and gate-based similarity provide complementary calibration information. 





\begin{table}[t]
\centering

\resizebox{\columnwidth}{!}{%
\begin{tabular}{lcccc}
\toprule
 & $h=3$ & $h=12$ & $h=24$ & $h=48$ \\
\midrule
\multicolumn{5}{c}{\textbf{Point Forecast (MSE)}} \\
\cmidrule{2-5}
Best baseline
& 0.718 (0.003) & 0.419 (0.013) & 0.395 (0.021) & 1.128 (0.047) \\
ABF-PatchTST
& 0.716 (0.003) & 0.420 (0.010) & 0.390 (0.014) & 1.125 (0.042) \\
ABF-T-PatchTST
& \textbf{0.715 (0.004)} & \textbf{0.417 (0.008)}
& \textbf{0.388 (0.011)} & \textbf{1.121 (0.046)} \\
\midrule
\multicolumn{5}{c}{\textbf{Uncertainty Quantification (IS)}} \\
\cmidrule{2-5}
Best baseline
& 3.675 (0.015) & 2.573 (0.024) & 2.444 (0.034) & 3.534 (0.053) \\
Time only
& 3.652 (0.012) & 2.550 (0.023) & 2.353 (0.026) & 3.553 (0.039) \\
Gate only
& 3.826 (0.152) & 3.306 (0.862) & 3.390 (0.991) & 6.215 (2.292) \\
Full asym. GLCP
& \textbf{3.557 (0.012)} & \textbf{2.463 (0.141)}
& \textbf{2.177 (0.213)} & \textbf{2.931 (0.430)} \\
\bottomrule
\end{tabular}%
}
\caption{PatchTST backbone modularity and GLCP ablations. Entries report mean MSE for point forecasting and mean IS for uncertainty quantification over 20 seeds, with standard deviations in parentheses.}
\label{tab:patchtst_modularity_glcp}
\end{table}

\paragraph{Beyond financial forecasting: Solar forecasting.}
Table~\ref{tab:solar_results} evaluates the proposed framework on a benchmark solar forecasting dataset \citep{godahewa2020solar}. ABF-T achieves competitive point forecasting accuracy, while GLCP consistently improves interval scores across all horizons. This demonstrates that the framework generalizes beyond the motivating financial forecasting application.
\begin{table}[!htbp]
\centering

\setlength{\tabcolsep}{4pt}
\begin{tabular}{c cc ccc}
\toprule
\multirow{2}{*}{$h$}
& \multicolumn{2}{c}{Point Forecast (MSE)}
& \multicolumn{3}{c}{Uncertainty Quantification} \\
\cmidrule(lr){2-3}
\cmidrule(lr){4-6}
& Best Baseline & ABF-T
& GLCP IS & Best IS & Gain \\
\midrule
3
& \textbf{0.154}
& \textbf{0.154}
& \textbf{1.943}
& 2.018
& +3.72\% \\

12
& \textbf{0.153}
& \textbf{0.153}
& \textbf{1.979}
& 2.037
& +2.83\% \\

24
& \textbf{0.128}
& 0.129
& \textbf{1.712}
& 1.797
& +4.70\% \\

48
& \textbf{0.164}
& \textbf{0.164}
& \textbf{1.549}
& 1.691
& +8.41\% \\

96
& \textbf{0.136}
& 0.138
& \textbf{1.513}
& 1.613
& +6.16\% \\

\bottomrule
\end{tabular}
\caption{Solar forecasting results. Left: point forecasting performance. Right: uncertainty quantification performance. }
\label{tab:solar_results}
\end{table}

\section{Conclusion and Discussion}

We proposed ABF-T-GLCP, a model-agnostic framework for adaptive forecasting
and uncertainty quantification in nonstationary multivariate time series. The
central idea is to couple point forecasting and conformal calibration through a
shared learned predictive-state representation: the forecasting gate adapts
multi-scale prediction while also identifying temporally recent and
regime-relevant calibration samples. This reuse of the gate representation
allows point forecasts and prediction intervals to adapt consistently under
evolving temporal dynamics.

Experiments on high-frequency commodity forecasting show consistent gains in
point forecasting accuracy and interval efficiency across multiple horizons.
Additional results on solar forecasting and alternative backbones indicate that
the framework extends beyond the motivating financial application and is not
tied to a particular base forecaster. The theoretical analysis gives an
approximate local coverage guarantee under predictable localization and local
score stability.

The framework relies on informative predictive-state representations. If the
gate fails to distinguish regimes, localized calibration may be less effective.
The additional computational cost mainly comes from training multiple
scale-specific experts, fitting the gate, and computing localized conformal
weights. Future work includes improving the efficiency of these components,
extending guarantees to broader forms of nonstationarity, and integrating
stronger domain-specific or foundation forecasting backbones.

\section*{Acknowledgments}
This research was supported by King Abdullah University of Science and Technology (KAUST).

\section*{Disclosure statement}\label{disclosure-statement}
The authors declare that they have no known competing financial
interests or personal relationships that could have appeared to
influence the work reported in this paper.

\newpage
\bibliographystyle{plainnat}
\bibliography{paper}

\newpage
\section*{Supplementary Material}

We provide additional details for the adaptive multi-scale
forecasting and Gate-Localized Conformal Prediction framework.


\subsection{Proof of Theorem~1}

Fix a forecasting horizon $h$ and test pair $(t,j)$, and condition on
$\mathcal F_t$. By the predictable-localization part of Assumption~1, the
weights $\{w_g^{(h,t,j)}:g\in\mathcal H_t\}$ are fixed under this
conditioning. Define the localized population CDFs for the lower- and
upper-tail scores as
\[
\begin{aligned}
\bar F_{t,j}^{-,(h)}(s)
&=
\sum_{g\in\mathcal H_t}
\bar w_g^{(h,t,j)}
F_{g,j}^{-,(h)}(s), \\
\bar F_{t,j}^{+,(h)}(s)
&=
\sum_{g\in\mathcal H_t}
\bar w_g^{(h,t,j)}
F_{g,j}^{+,(h)}(s).
\end{aligned}
\]
By the definition of the weighted empirical quantiles used in Algorithm~1,
\[
\widehat F_{t,j}^{-,w,(h)}(q_{t,j}^{-,(h)})
\ge
1-\alpha/2,
\qquad
\widehat F_{t,j}^{+,w,(h)}(q_{t,j}^{+,(h)})
\ge
1-\alpha/2.
\]
On the joint weighted empirical concentration event,
\[
\bar F_{t,j}^{-,(h)}(q_{t,j}^{-,(h)})
\ge
1-\alpha/2-\epsilon_{t,j}^{-,(h)}(\eta),
\]
and
\[
\bar F_{t,j}^{+,(h)}(q_{t,j}^{+,(h)})
\ge
1-\alpha/2-\epsilon_{t,j}^{+,(h)}(\eta).
\]
By the tail-wise local-stability part of Assumption~1, the localized
population CDFs approximate the corresponding test score distributions up to
errors $\delta_{t,j}^{-,(h)}$ and $\delta_{t,j}^{+,(h)}$. Therefore,
\[
F_{t,j}^{-,(h)}(q_{t,j}^{-,(h)})
\ge
1-\alpha/2
-\epsilon_{t,j}^{-,(h)}(\eta)
-\delta_{t,j}^{-,(h)},
\]
and
\[
F_{t,j}^{+,(h)}(q_{t,j}^{+,(h)})
\ge
1-\alpha/2
-\epsilon_{t,j}^{+,(h)}(\eta)
-\delta_{t,j}^{+,(h)}.
\]
Equivalently,
\[
\Pr
\left(
S_{t,j}^{-,(h)}>q_{t,j}^{-,(h)}
\mid \mathcal F_t
\right)
\le
\alpha/2+\delta_{t,j}^{-,(h)}
+\epsilon_{t,j}^{-,(h)}(\eta),
\]
and
\[
\Pr
\left(
S_{t,j}^{+,(h)}>q_{t,j}^{+,(h)}
\mid \mathcal F_t
\right)
\le
\alpha/2+\delta_{t,j}^{+,(h)}
+\epsilon_{t,j}^{+,(h)}(\eta).
\]

The interval produced by Algorithm~1 is
\[
C_{t,j}^{(h),1-\alpha}
=
\left[
\hat y^{\mathrm{ABF\text{-}T}}_{t+h,j}
-
q_{t,j}^{-,(h)}\hat\sigma_j^{(h)},
\;
\hat y^{\mathrm{ABF\text{-}T}}_{t+h,j}
+
q_{t,j}^{+,(h)}\hat\sigma_j^{(h)}
\right].
\]
By the definitions
\[
S_{t,j}^{-,(h)}=\max\{-E_{t,j}^{(h)},0\},
\qquad
S_{t,j}^{+,(h)}=\max\{E_{t,j}^{(h)},0\},
\]
coverage can fail only if the lower-tail score exceeds its calibrated radius
or the upper-tail score exceeds its calibrated radius:
\[
\{y_{t+h,j}\notin C_{t,j}^{(h),1-\alpha}\}
\subseteq
\{S_{t,j}^{-,(h)}>q_{t,j}^{-,(h)}\}
\cup
\{S_{t,j}^{+,(h)}>q_{t,j}^{+,(h)}\}.
\]
Applying the union bound gives
\[
\begin{aligned}
&\Pr
\left(
y_{t+h,j}\notin C_{t,j}^{(h),1-\alpha}
\mid \mathcal F_t
\right) \\
&\le
\alpha
+\delta_{t,j}^{-,(h)}
+\delta_{t,j}^{+,(h)}
+\epsilon_{t,j}^{-,(h)}(\eta)
+\epsilon_{t,j}^{+,(h)}(\eta).
\end{aligned}
\]
Rearranging yields
\[
\begin{aligned}
&\Pr
\left(
y_{t+h,j}\in C_{t,j}^{(h),1-\alpha}
\mid \mathcal F_t
\right) \\
&\ge
1-\alpha
-\delta_{t,j}^{-,(h)}
-\delta_{t,j}^{+,(h)}
-\epsilon_{t,j}^{-,(h)}(\eta)
-\epsilon_{t,j}^{+,(h)}(\eta).
\end{aligned}
\]
Since the joint weighted empirical concentration event holds with probability
at least $1-\eta$ over the calibration history, the result follows.

\subsection{Computational Complexity}

Let $N$ be the number of series, $T$ the number of time points, $K$ the number
of temporal scales, and $d$ the dimension of the gate/regime features. The
scale-specific forecasting experts are trained independently, with total cost
\[
\sum_{k=1}^K \mathcal O\!\left(C_{\mathrm{fit}}(L_k,N,T)\right),
\]
where $C_{\mathrm{fit}}(L_k,N,T)$ depends on the chosen base forecaster and
lookback length $L_k$. The gating network evaluates $K$ scale weights for each
series and time point, giving cost $\mathcal O(TNKd)$ up to network constants.
The validation-based anchor and blending selection costs $\mathcal O(TNK)$.

The sparse transfer step fits one regularized residual model per target series.
If $p=\mathcal O(NK)$ denotes the number of gate-weighted source-scale
features, its cost is approximately
\[
\mathcal O\!\left(N\,C_{\mathrm{reg}}(T_{\mathrm{val}},p)\right),
\]
where $C_{\mathrm{reg}}$ is the cost of the chosen regularized regression
solver. In practice, this step is controlled by validation selection and sparse
regularization.

For GLCP, constructing intervals requires computing localized weights over a
rolling calibration window of size $M$. For each test time and target series,
this costs $\mathcal O(MK)$ for gate-similarity weights and
$\mathcal O(M\log M)$ for the weighted quantile. Thus the online calibration
cost over $T_{\mathrm{test}}$ test points is
\[
\mathcal O\!\left(T_{\mathrm{test}}N\{MK+M\log M\}\right).
\]
Overall, for fixed $K$ and calibration window size $M$, the online calibration
stage scales linearly in the number of series and test points. The main
additional overhead relative to standard conformal calibration comes from
gate-similarity weighting and sparse predictive transfer.

\subsection{The complete results for the financial commodity data}
Here, we provide the full details for the experiment we conducted on the financial commodity data. 

\paragraph{Computing infrastructure.}
Experiments were conducted on a Linux-based high-performance computing cluster running Rocky Linux 9.4 (Blue Onyx). The main HAR-based commodity experiments were run primarily on CPU compute nodes, with each array task allocated 8 CPU cores and 64\,GB RAM on nodes equipped with Intel Xeon Gold 6248 CPUs (2.50\,GHz). Neural-backbone ablations, including PatchTST, were run on GPU compute nodes with one GPU allocated per array task, together with 8 CPU cores and 96\,GB RAM. The implementation was developed in Python 3.11.0 using NumPy 1.26.2, pandas 2.3.3, scikit-learn 1.7.2, PyTorch 2.5.1 with CUDA 12.1 support, Matplotlib 3.8.2, and PyArrow 24.0.0. All experiments used chronological train--validation--calibration--test splits, and all model-selection, transfer-selection, and conformal-calibration steps were performed without using the test set.

\subsection{Hyperparameter Selection and Reproducibility Details}
\label{app:hyperparameters}

To avoid information leakage, all experiments use a strict chronological split
\[
\textsc{Train} \rightarrow \textsc{Val} \rightarrow \textsc{Cal} \rightarrow \textsc{Test}.
\]
The split ratios are fixed as \(50\%\), \(15\%\), \(20\%\), and \(15\%\), respectively. The training set is used to estimate forecasting experts and the gating network. The validation set is used for all forecasting model selection and hyperparameter tuning. The calibration set is used only for conformal calibration, and the test set is used only for final evaluation.

\paragraph{Forecasting horizons and scale library.}
We evaluate horizons
\[
h \in \{3,12,24,48,96\}
\]
using 5-minute bars, corresponding to 15, 60, 120, 240, and 480 minutes ahead. The multi-scale expert library uses fixed lookback windows
\[
\mathcal L=\{1,3,6,12,24,48,96,240\}.
\]
These lookback values are fixed across all horizons and seeds.

\paragraph{Forecasting experts.}
Each scale-specific forecasting expert \(f_k\) is trained on the training set. For the HAR expert backbone, ridge regression is used with ridge penalty
\[
\lambda_{\mathrm{ridge}} = 10.
\]
No test data are used in fitting any expert.

\paragraph{Asset-specific gating network.}
The gating network \(g_\theta(z_{t,j})\) maps asset-specific regime features to scale weights
\[
\pi_{t,j}=g_\theta(z_{t,j}), \qquad \sum_{k=1}^K \pi_{t,j,k}=1.
\]
The gate is trained on the training set with hidden dimension \(64\), dropout \(0.10\), learning rate \(10^{-3}\), weight decay \(10^{-4}\), batch size \(512\), and \(200\) training epochs. The softmax temperature is fixed to \(0.80\), and top-\(k\) gating uses \(k=3\). The gate regularization coefficients are fixed as
\[
\lambda_{\mathrm{ent}}=0.0002,\qquad
\lambda_{\mathrm{bal}}=0.0002,\qquad
\lambda_{\mathrm{usage}}=0.0005.
\]

\paragraph{ABF anchor selection and multi-scale correction.}
For each horizon and seed, ABF first selects the validation-best forecasting expert as the anchor:
\[
a
=
\arg\min_{k}
\mathrm{MSE}_{\mathrm{val}}(\hat y_{\cdot,\cdot,k}).
\]
The ABF forecast is then
\[
\hat y^{\mathrm{ABF}}_{t,j}
=
\hat y_{t,j,a}
+
\rho
\sum_{k=1}^K
\pi_{t,j,k}
\left(
\hat y_{t,j,k}
-
\hat y_{t,j,a}
\right).
\]
The correction strength is selected on the validation set from
\[
\rho \in
\{0,0.025,0.05,0.075,0.10,0.15,0.20,0.30,0.40,0.50\},
\]
using validation MSE. Thus, \(\rho=0\) recovers the anchor expert, and positive values are used only when multi-scale corrections improve validation performance.

\paragraph{Sparse predictive transfer.}
After ABF, we model residual spillovers using
\[
e_{t,j}
=
y_{t,j}
-
\hat y^{\mathrm{ABF}}_{t,j},
\]
and construct the transfer feature vector
\[
\phi_t
=
\mathrm{vec}
\left(
\{\pi_{t,i,k}\hat y_{t,i,k}\}_{i=1,\ldots,N;\,k=1,\ldots,K}
\right).
\]
For each target series \(j\), the residual correction is modeled as
\[
e_{t,j}
\approx
b_{0,j}
+
\phi_t^\top b_j.
\]
The transfer hyperparameters are selected using an inner split of the validation set: \(70\%\) of the validation set is used to fit candidate transfer models, and the remaining \(30\%\) is used to select the configuration by validation MSE.

We evaluate the following transfer variants: 
\begin{itemize} 
\item Lasso with intercept;
\item Lasso without intercept;
\item Elastic Net with intercept;
\item Elastic Net without intercept;
\item Ridge without intercept. 
\end{itemize}
The Lasso penalty grid is
\[
\lambda_1 \in
\{0.001,0.003,0.01,0.03,0.05,0.10\},
\]
the Ridge penalty grid is
\[
\lambda_2 \in
\{0.1,1,10,100\},
\]
and the Elastic Net mixing ratio grid is
\[
r \in \{0.15,0.50,0.85\}.
\]
The transfer shrinkage parameter is selected from
\[
\gamma \in \{0,0.25,0.50,0.75,1.0\}.
\]
To avoid negative transfer, the selected transfer model is accepted only if it improves the inner-validation MSE by at least \(1\%\) relative to the no-transfer ABF model. Otherwise, the method falls back to \(\gamma=0\), i.e., no transfer correction.

\paragraph{Conformal calibration parameters.}
All conformal methods use target coverage \(1-\alpha=0.90\), with \(\alpha=0.10\). The rolling calibration window is fixed to
\[
W=2000.
\]
The calibration set is used only to construct conformal residual distributions or select conformal calibration parameters; it is never used to fit the forecasting model.

\paragraph{GLCP hyperparameters.}
GLCP assigns each historical calibration observation \(g\in\mathcal H_t\) the weight
\[
w^{(t,j)}_g
=
\lambda^{t-g}
\exp\left(
-
\frac{
\|\pi_{g,j}-\pi_{t,j}\|_2^2
}{
2\tau^2
}
\right),
\]
where \(\lambda\) controls temporal recency and \(\tau\) controls gate-based localization. We fix
\[
\lambda=0.995.
\]
The localization bandwidth is selected from
\[
\tau \in
\{0.10,0.25,0.50,0.75,1.00,1.50,2.00\}.
\]
The selected \(\tau\) minimizes the conformal interval score subject to empirical coverage being within \(0.5\%\) of the nominal target on the validation-calibration selection split. After selection, the chosen \(\tau\) is fixed for final test evaluation.

\paragraph{Random seeds and reporting.}
All main experiments are repeated over 20 random seeds:
\[
\begin{aligned}
\{113, 307, 401, 509, 601, 709, 809, 907, 1009, 1201,\\
1303, 1409, 1511, 1601, 1709, 1801, 1907, 2003, 2111, 2203\}.
\end{aligned}
\]
All reported results are means and standard deviations across these seeds. No hyperparameter is selected using the test set.

\subsection{Analysis of point forecasting results}
The detailed results in Table~\ref{tab:full_point_forecast} provide further insight into the design choices underlying ABF-T.

First, the single-scale HAR experts exhibit substantial variation across
forecasting horizons, indicating that no single temporal scale is uniformly
optimal. The best-performing lookback changes with the prediction horizon,
making it impossible to determine a priori which expert should be trusted for a
future forecasting task. This naturally motivates adaptive multi-scale
forecasting through a mixture-of-experts framework.

However, the results also show that naively combining multiple experts is not
always beneficial. Although the gated mixture adaptively assigns expert
weights, it frequently performs worse than the strongest single-scale expert.
This suggests that including weaker experts may introduce unnecessary
estimation variance and degrade point forecasting accuracy, even when the gate
is learned adaptively.

The proposed Adaptive Best Forecasting (ABF) is designed to address this
trade-off. Rather than averaging all experts indiscriminately, ABF first
identifies the validation-best expert as a reliable forecasting anchor and
allows the remaining experts to contribute only through validation-supported
gated corrections. Consequently, ABF without predictive transfer consistently
matches or improves upon the strongest single-scale expert while avoiding the
performance degradation occasionally observed in conventional
mixture-of-experts approaches.

Finally, Table~\ref{tab:full_point_forecast} reports several predictive-transfer
variants as ablation studies. These variants isolate the contribution of the
second stage of the framework after the adaptive forecasting component has been
constructed. Compared with ABF without transfer, sparse predictive transfer
consistently provides additional improvements, indicating that residual
cross-series predictive information remains after the dominant temporal
dynamics have been captured. Different regularization schemes produce similar
performance, while the final ABF-T model adopts the transfer configuration
selected exclusively on the validation set. This validation-driven selection
maintains the model-agnostic nature of the framework and helps exploit useful
cross-series dependencies without introducing negative transfer.

Overall, the results demonstrate that the proposed framework addresses two
complementary challenges: ABF resolves the pooling--specialization trade-off in
adaptive multi-scale forecasting, while sparse predictive transfer further
leverages residual cross-series information to improve forecasting accuracy,
particularly at longer forecasting horizons.

\begin{table*}[t]
\centering

\scriptsize
\setlength{\tabcolsep}{3.0pt}
\resizebox{\textwidth}{!}{%
\begin{tabular}{lcc cc cc cc cc}
\toprule
\multirow{2}{*}{Method} & \multicolumn{2}{c}{$h=3$} & \multicolumn{2}{c}{$h=12$} & \multicolumn{2}{c}{$h=24$} & \multicolumn{2}{c}{$h=48$} & \multicolumn{2}{c}{$h=96$} \\
\cmidrule(lr){2-3}\cmidrule(lr){4-5}\cmidrule(lr){6-7}\cmidrule(lr){8-9}\cmidrule(lr){10-11}
& MSE & MAE & MSE & MAE & MSE & MAE & MSE & MAE & MSE & MAE \\
\midrule
\multicolumn{11}{l}{\textbf{Single-scale HAR experts}}\\
Single-scale HAR ($L=1$) & 0.862(0.000) & 0.661(0.000) & 0.628(0.000) & 0.599(0.000) & 0.963(0.000) & 0.779(0.000) & 1.946(0.000) & 1.105(0.000) & 3.848(0.000) & 1.528(0.000) \\
Single-scale HAR ($L=3$) & 0.863(0.000) & 0.662(0.000) & 0.643(0.000) & 0.608(0.000) & 0.926(0.000) & 0.760(0.000) & 1.987(0.000) & 1.116(0.000) & 3.915(0.000) & 1.542(0.000) \\
Single-scale HAR ($L=6$) & 0.870(0.000) & 0.664(0.000) & 0.677(0.000) & 0.626(0.000) & 0.908(0.000) & 0.748(0.000) & 2.060(0.000) & 1.135(0.000) & 4.064(0.000) & 1.574(0.000) \\
Single-scale HAR ($L=12$) & 0.890(0.000) & 0.672(0.000) & 0.804(0.000) & 0.675(0.000) & 0.945(0.000) & 0.759(0.000) & 2.214(0.000) & 1.178(0.000) & 4.424(0.000) & 1.647(0.000) \\
Single-scale HAR ($L=24$) & 0.964(0.000) & 0.703(0.000) & 0.923(0.000) & 0.706(0.000) & 0.883(0.000) & 0.721(0.000) & 2.541(0.000) & 1.266(0.000) & 5.370(0.000) & 1.826(0.000) \\
Single-scale HAR ($L=48$) & 1.107(0.000) & 0.765(0.000) & 1.144(0.000) & 0.801(0.000) & 1.300(0.000) & 0.889(0.000) & 3.323(0.000) & 1.445(0.000) & 6.712(0.000) & 2.060(0.000) \\
Single-scale HAR ($L=96$) & 1.619(0.000) & 0.959(0.000) & 2.996(0.000) & 1.340(0.000) & 4.459(0.000) & 1.716(0.000) & 7.784(0.000) & 2.200(0.000) & 6.861(0.000) & 2.090(0.000) \\
Single-scale HAR ($L=240$) & 4.469(0.000) & 1.707(0.000) & 7.814(0.000) & 2.208(0.000) & 13.207(0.000) & 2.904(0.000) & 24.418(0.000) & 3.891(0.000) & 12.978(0.000) & 2.906(0.000) \\
\midrule
\multicolumn{11}{l}{\textbf{Multi-scale HAR baselines}}\\
Equal-weight multi-scale HAR & 1.005(0.000) & 0.735(0.000) & 1.013(0.000) & 0.767(0.000) & 1.229(0.000) & 0.866(0.000) & 3.355(0.000) & 1.442(0.000) & 4.787(0.000) & 1.724(0.000) \\
Gated multi-scale HAR & 1.290(0.221) & 0.845(0.096) & 1.977(0.772) & 1.050(0.209) & 2.621(0.879) & 1.254(0.224) & 6.560(2.187) & 1.969(0.349) & 6.366(0.650) & 2.013(0.115) \\
Anchored best-scale blend & 0.872(0.003) & 0.664(0.001) & 0.629(0.001) & 0.599(0.001) & 0.925(0.008) & 0.759(0.004) & 1.949(0.007) & 1.106(0.002) & 3.848(0.000) & 1.528(0.000) \\
Anchor-safe multi-scale gate & 0.872(0.003) & 0.664(0.001) & 0.629(0.001) & 0.599(0.001) & 0.925(0.007) & 0.759(0.003) & 1.948(0.004) & 1.105(0.001) & 3.848(0.000) & 1.528(0.000) \\
Anchor-regularized gate & 1.320(0.231) & 0.861(0.097) & 1.959(0.576) & 1.059(0.163) & 2.760(0.886) & 1.292(0.213) & 6.873(2.083) & 2.017(0.325) & 6.506(0.756) & 2.036(0.134) \\
\midrule
\multicolumn{11}{l}{\textbf{ABF-T variants}}\\
ABF-T & 0.851(0.009) & 0.655(0.004) & 0.601(0.019) & 0.584(0.009) & 0.828(0.237) & 0.687(0.046) & 1.752(0.117) & 1.043(0.034) & 3.270(0.207) & 1.410(0.046) \\
ABF-T w/o transfer & 0.872(0.003) & 0.664(0.001) & 0.629(0.001) & 0.599(0.001) & 0.925(0.007) & 0.759(0.003) & 1.948(0.004) & 1.105(0.001) & 3.848(0.000) & 1.528(0.000) \\
ABF-T selected transfer & \textbf{0.849(0.009)} & \textbf{0.654(0.004)} & 0.599(0.019) & 0.583(0.009) & 0.795(0.203) & 0.679(0.037) & 1.683(0.070) & 1.022(0.021) & 3.161(0.206) & 1.383(0.040) \\
ABF-T Lasso & 0.850(0.010) & 0.655(0.004) & \textbf{0.598(0.015)} & \textbf{0.583(0.008)} & \textbf{0.750(0.072)} & \textbf{0.677(0.034)} & \textbf{1.663(0.080)} & \textbf{1.016(0.024)} & 3.175(0.229) & 1.387(0.045) \\
ABF-T Lasso no intercept & 0.865(0.010) & 0.661(0.005) & 0.610(0.015) & 0.589(0.008) & 0.847(0.088) & 0.720(0.029) & 1.771(0.098) & 1.052(0.028) & 3.499(0.256) & 1.469(0.050) \\
ABF-T ElasticNet & 0.850(0.010) & 0.655(0.004) & 0.599(0.019) & 0.583(0.009) & 0.795(0.202) & 0.679(0.037) & 1.687(0.064) & 1.024(0.019) & \textbf{3.155(0.205)} & \textbf{1.381(0.040)} \\
ABF-T ElasticNet no intercept & 0.863(0.010) & 0.660(0.005) & 0.607(0.017) & 0.588(0.009) & 0.877(0.185) & 0.720(0.033) & 1.793(0.095) & 1.059(0.027) & 3.477(0.249) & 1.464(0.050) \\
ABF-T Ridge no intercept & 0.872(0.003) & 0.664(0.001) & 0.628(0.002) & 0.599(0.001) & 1.465(2.309) & 0.787(0.136) & 1.965(0.103) & 1.105(0.028) & 3.486(0.147) & 1.460(0.030) \\
\bottomrule
\end{tabular}%
}
\caption{Complete point forecasting results on the commodity dataset. Each entry reports mean standard deviation (in parentheses) across repeated runs. Lower values are better; bold indicates the best result for each horizon and metric.}
\label{tab:full_point_forecast}
\end{table*}

\subsection{Analysis of conformal prediction results}

The main paper reports the average uncertainty quantification performance of all
methods, together with the coverage--interval score trade-off and a visual
comparison of prediction intervals for the representative forecasting horizon
$h=48$. In this supplementary section, we provide additional analyses that
complement the main results. Specifically, we report the variability of each
method across repeated runs, visualize the coverage--width trade-off, and
present prediction interval examples for the remaining forecasting horizons
($h=3$, $12$, $24$, and $96$).


\begin{table*}[t]
\centering
\scriptsize
\setlength{\tabcolsep}{2.2pt}
\resizebox{\textwidth}{!}{%
\begin{tabular}{l ccc ccc ccc ccc ccc}
\toprule
\multirow{2}{*}{Method}
& \multicolumn{3}{c}{$h=3$}
& \multicolumn{3}{c}{$h=12$}
& \multicolumn{3}{c}{$h=24$}
& \multicolumn{3}{c}{$h=48$}
& \multicolumn{3}{c}{$h=96$} \\
\cmidrule(lr){2-4}
\cmidrule(lr){5-7}
\cmidrule(lr){8-10}
\cmidrule(lr){11-13}
\cmidrule(lr){14-16}
& Cov & Width & IS
& Cov & Width & IS
& Cov & Width & IS
& Cov & Width & IS
& Cov & Width & IS \\
\midrule

ACI
& 0.000 & 0.02 & 0.02
& 0.000 & 0.05 & 0.06
& 0.001 & 0.14 & 0.31
& 0.001 & 0.15 & 0.18
& 0.002 & 0.13 & 0.17 \\

AgACI
& 0.000 & 0.02 & 0.02
& 0.001 & 0.05 & 0.06
& 0.001 & 0.14 & 0.31
& 0.001 & 0.17 & 0.20
& 0.002 & 0.13 & 0.18 \\

RLCP
& 0.004 & 0.07 & 0.05
& 0.004 & 0.25 & 0.23
& 0.005 & 0.23 & 0.38
& 0.004 & 0.62 & 0.63
& 0.004 & 0.49 & 0.51 \\

RCQR
& 0.003 & 0.09 & 0.07
& 0.003 & 0.36 & 0.35
& 0.005 & 0.42 & 0.44
& 0.003 & 0.47 & 0.47
& 0.003 & 0.39 & 0.44 \\

EnbPI
& 0.001 & 0.05 & 0.05
& 0.001 & 0.12 & 0.11
& 0.001 & 0.14 & 0.24
& 0.001 & 0.65 & 0.66
& 0.002 & 0.37 & 0.37 \\

SAOCP
& 0.001 & 0.02 & 0.03
& 0.001 & 0.06 & 0.07
& 0.002 & 0.16 & 0.33
& 0.002 & 0.17 & 0.20
& 0.003 & 0.15 & 0.22 \\

MoECP
& 0.012 & 0.28 & 0.17
& 0.025 & 1.20 & 1.07
& 0.026 & 1.46 & 1.35
& 0.023 & 4.93 & 4.82
& 0.022 & 3.79 & 3.76 \\

\midrule
\textbf{GLCP}
& {0.001} & {0.02} & {0.01}
& {0.002} & {0.13} & {0.13}
& {0.003} & {0.18} & {0.31}
& {0.002} & {0.14} & {0.15}
& {0.003} & {0.08} & {0.11} \\

\bottomrule
\end{tabular}%
}
\caption{
Standard deviations of the uncertainty quantification metrics across repeated
runs. Cov, Width, and IS denote empirical coverage, average prediction interval
width, and interval score, respectively. Smaller standard deviations indicate
more stable performance across different random seeds.
}
\label{tab:uq_results_std}
\end{table*}

Table~\ref{tab:uq_results_std} summarizes the variability of each conformal
prediction method across repeated runs. Overall, GLCP exhibits consistently
small standard deviations for empirical coverage, interval width, and interval
score across all forecasting horizons, indicating that the proposed localized
calibration procedure is not only accurate but also stable with respect to
random initialization. Most competing methods also show relatively stable
coverage, while methods relying on stronger localization, such as RLCP and
MoECP, exhibit noticeably larger variability in interval width and interval
score, particularly for longer forecasting horizons.


\begin{figure*}[t]
    \centering
    \includegraphics[width=1\linewidth]{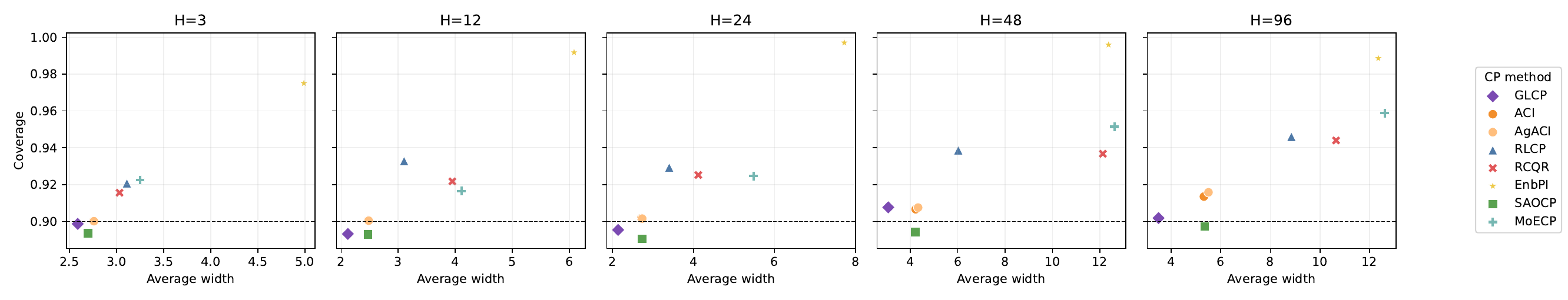}
    \caption{
    Coverage--width trade-off for all conformal prediction methods across
    forecasting horizons. Compared with the coverage--interval score comparison
    reported in the main paper, this figure directly illustrates the trade-off
    between calibration accuracy and interval sharpness.
    }
    \label{fig:coverage_width}
\end{figure*}

Figure~\ref{fig:coverage_width} complements the coverage--interval score
comparison presented in the main paper by directly comparing empirical coverage
against average prediction interval width. An ideal method is located near the
target coverage level while maintaining narrow prediction intervals. Across all
forecasting horizons, GLCP remains among the methods closest to this desirable
operating region, achieving competitive empirical coverage with consistently
sharper prediction intervals than the majority of competing approaches.


\begin{figure*}[t]
    \centering
    \includegraphics[width=1\linewidth]{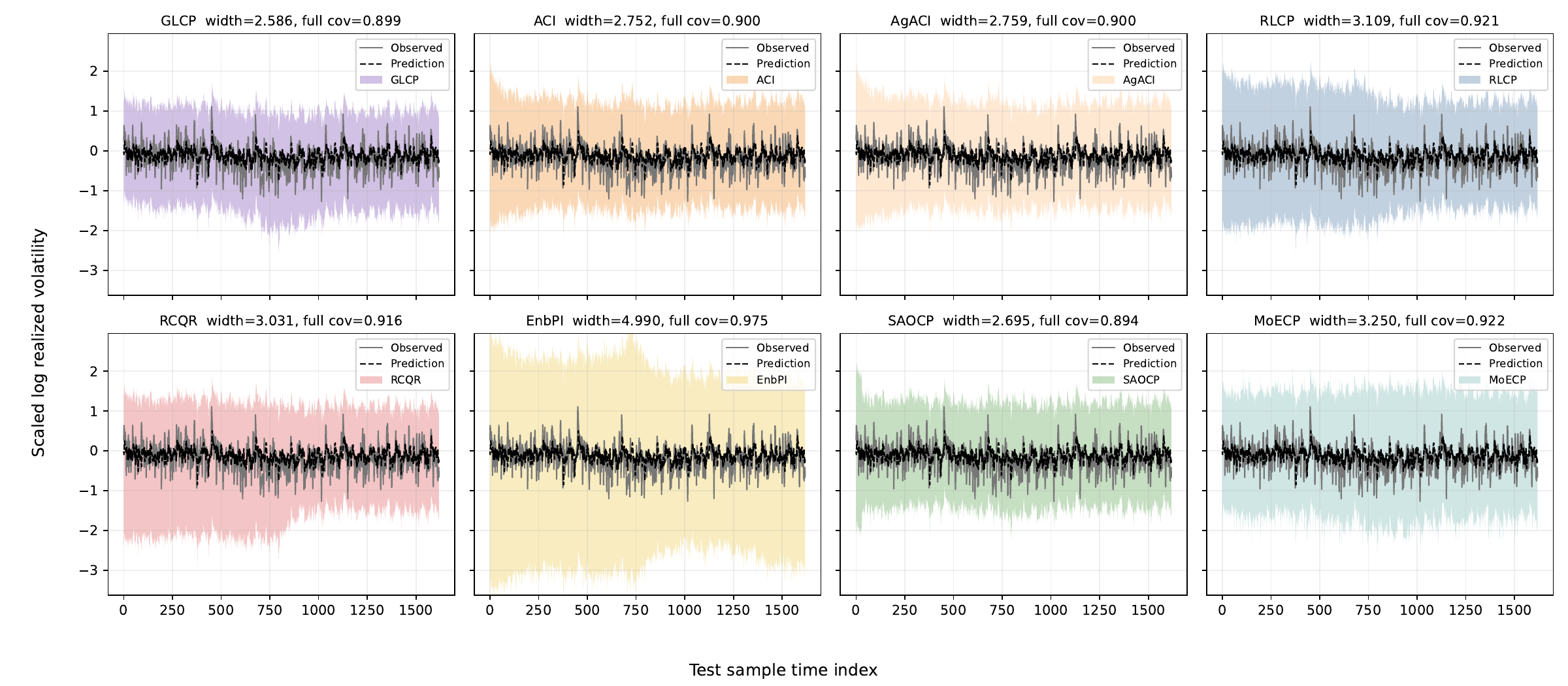}
    \caption{
    Prediction interval comparison for a representative segment of the test set
    at forecasting horizon $h=3$. Each panel shows the observed series, point
    forecasts, and the corresponding prediction intervals for one conformal
    prediction method.
    }
    \label{fig:interval_h3}
\end{figure*}

Figure~\ref{fig:interval_h3} illustrates the prediction intervals for the
shortest forecasting horizon. GLCP produces intervals that closely track the
observed series while remaining noticeably narrower than several competing
methods, demonstrating that localized calibration improves interval efficiency
without sacrificing empirical coverage.


\begin{figure*}[t]
    \centering
    \includegraphics[width=1\linewidth]{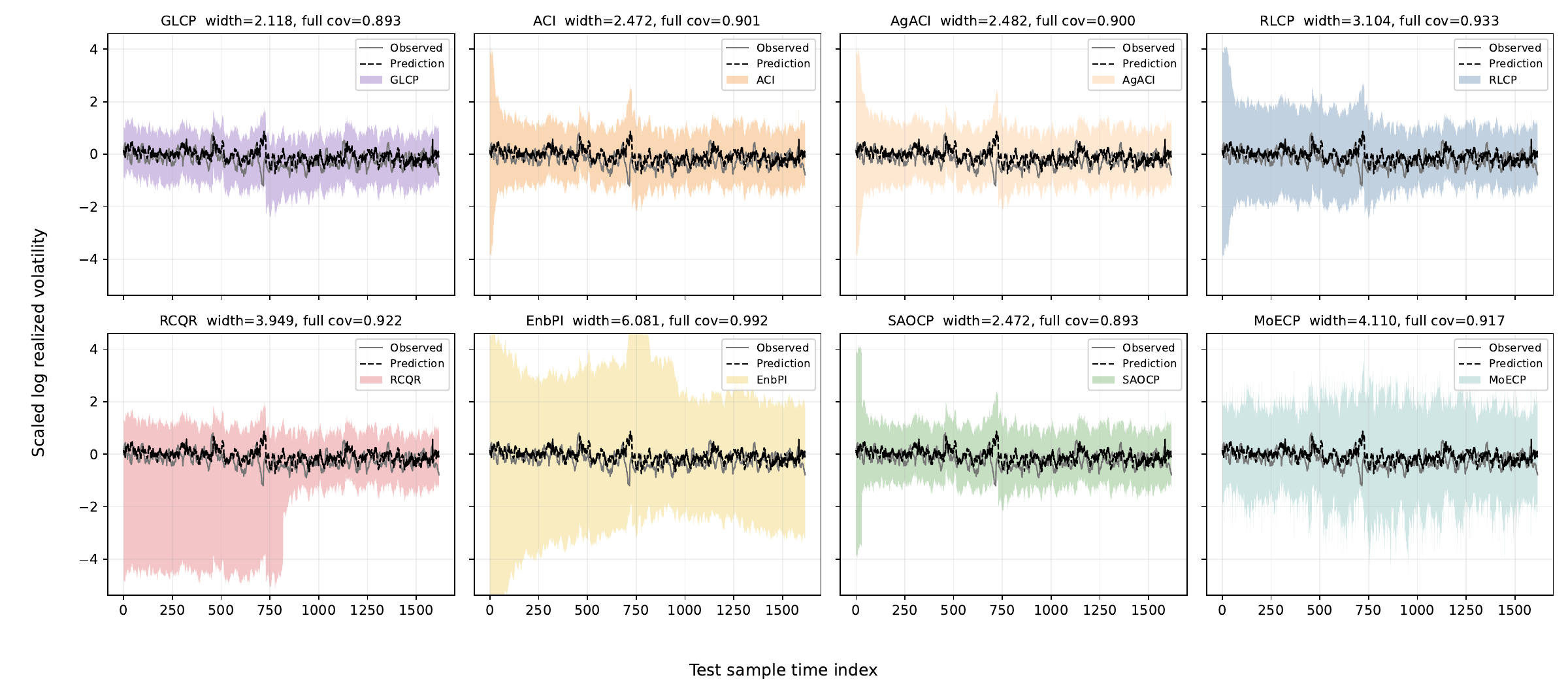}
    \caption{
    Prediction interval comparison for a representative segment of the test set
    at forecasting horizon $h=12$.
    }
    \label{fig:interval_h12}
\end{figure*}

Figure~\ref{fig:interval_h12} shows that the relative behavior of the competing
methods remains consistent at the one-hour forecasting horizon. GLCP continues
to maintain competitive coverage while producing comparatively compact
prediction intervals.


\begin{figure*}[t]
    \centering
    \includegraphics[width=1\linewidth]{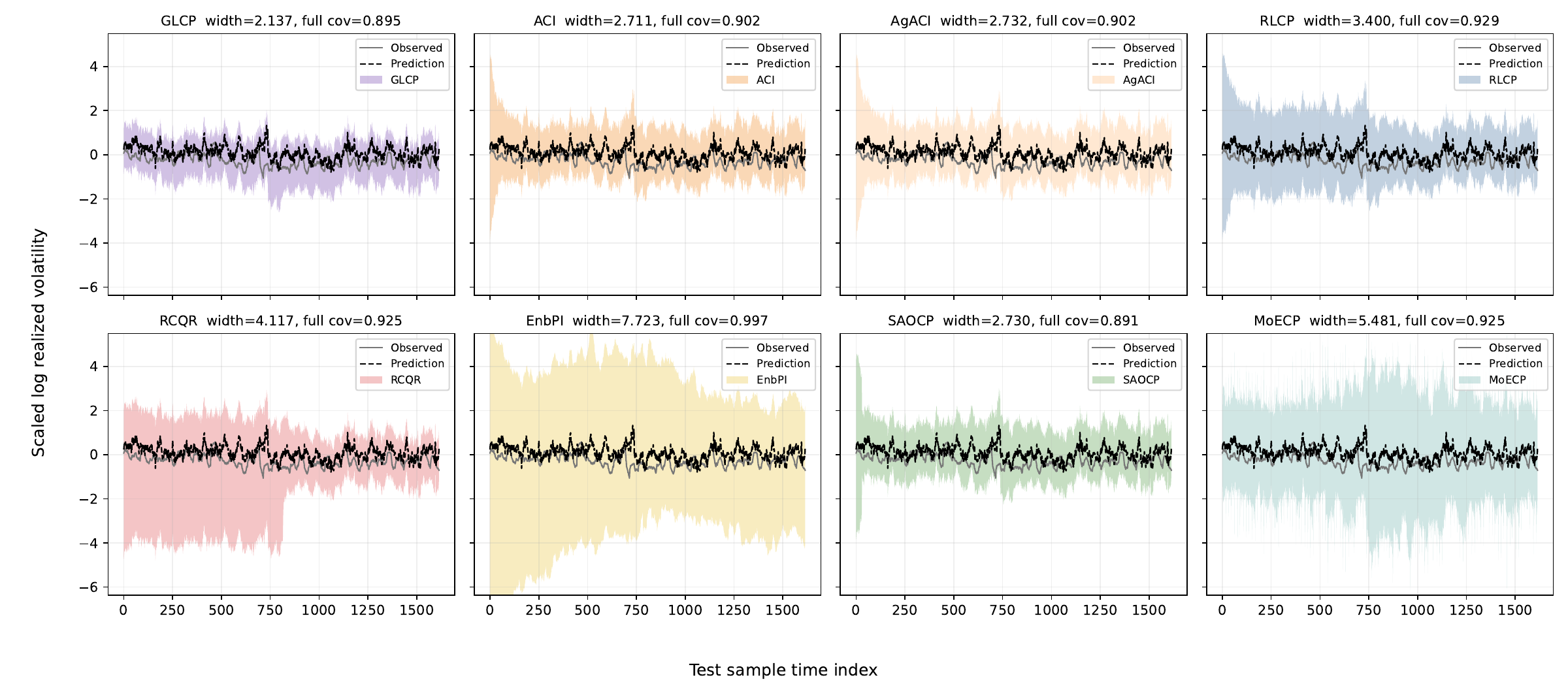}
    \caption{
    Prediction interval comparison for a representative segment of the test set
    at forecasting horizon $h=24$.
    }
    \label{fig:interval_h24}
\end{figure*}

Figure~\ref{fig:interval_h24} further demonstrates that the proposed
localization strategy remains effective for medium forecasting horizons. While
some competing methods substantially widen their prediction intervals to
maintain coverage, GLCP preserves relatively sharp intervals that continue to
follow the changing volatility dynamics.


\begin{figure*}[t]
    \centering
    \includegraphics[width=1\linewidth]{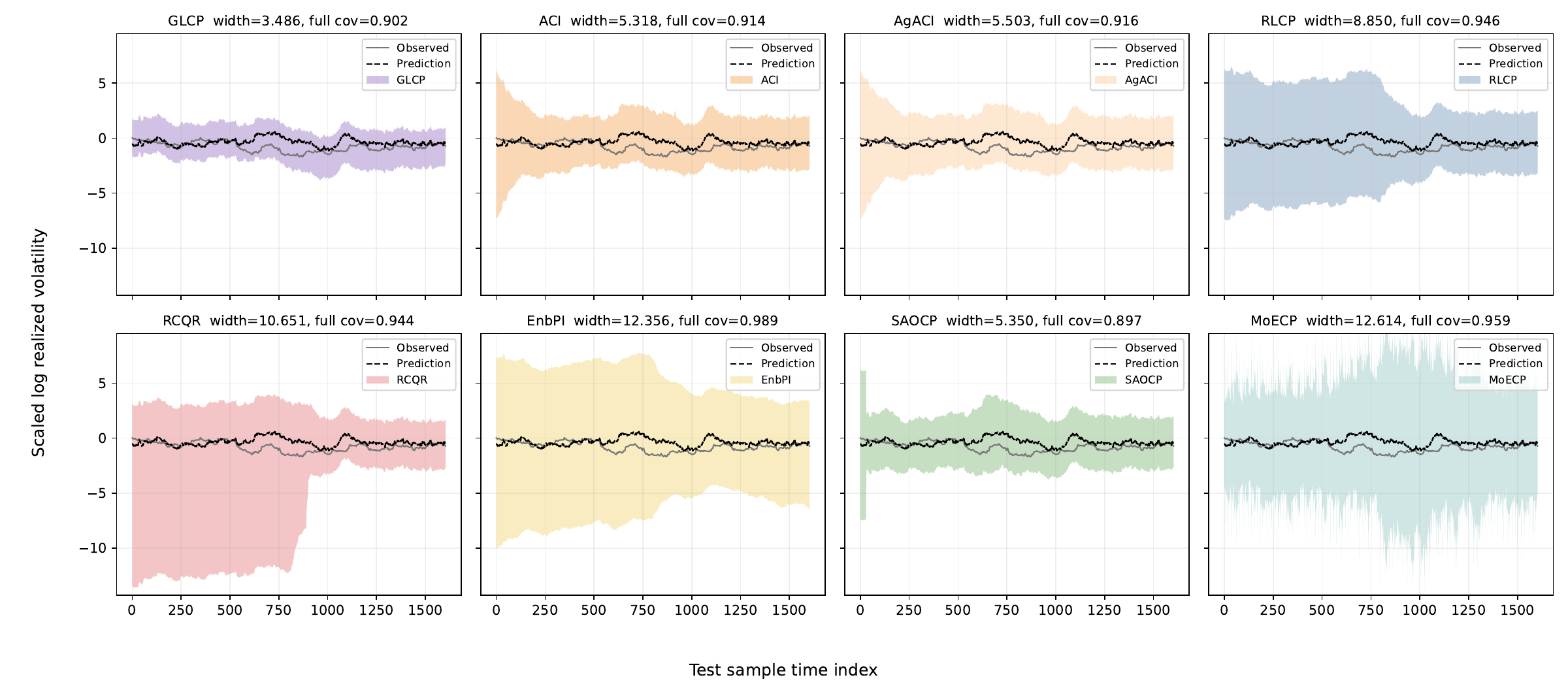}
    \caption{
    Prediction interval comparison for a representative segment of the test set
    at forecasting horizon $h=96$. 
    }
    \label{fig:interval_h96}
\end{figure*}

Figure~\ref{fig:interval_h96} illustrates the most challenging long-horizon
forecasting scenario considered in this work. Despite the increased uncertainty
associated with long-range prediction, GLCP continues to generate intervals
that are substantially sharper than many competing approaches while maintaining
coverage close to the target level. Together with the $h=48$ visualization in
the main paper, these additional examples demonstrate that the qualitative
advantages of GLCP are consistent across the full range of forecasting
horizons.

\subsection{Asset-specific predictive heterogeneity}

A central motivation of Adaptive Best Forecasting (ABF) is that predictive
heterogeneity exists along two dimensions. First, different commodities exhibit
distinct temporal dynamics because they are influenced by different market
mechanisms, sector-specific factors, and trading behaviors. Second, the
predictive structure of a single commodity is itself nonstationary, so the most
informative temporal scale evolves over time as market conditions change.
Consequently, there is generally no single lookback window that remains optimal
throughout the forecasting period. This motivates learning asset-specific,
time-varying temporal representations rather than relying on a fixed forecasting
scale.

\begin{figure*}[t]
    \centering
    \includegraphics[width=\linewidth]{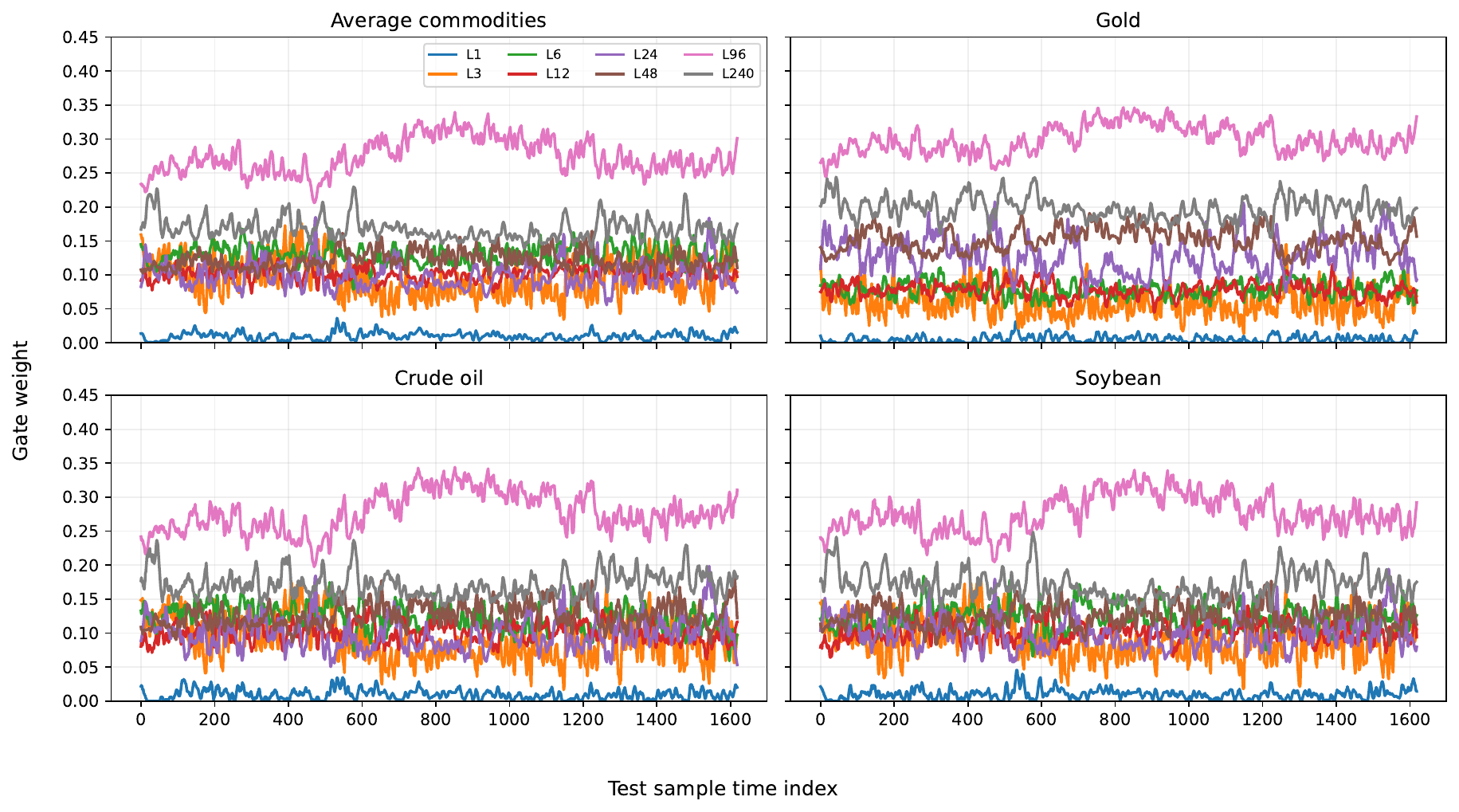}
    \caption{
    Learned asset-specific gate dynamics for forecasting horizon $h=3$.
    The first panel reports the average gate weights across all commodities,
    while the remaining panels show representative commodities (gold, crude
    oil, and soybean). The gate continuously reallocates importance among
    temporal experts, illustrating both commodity-specific and time-varying
    forecasting regimes.
    }
    \label{fig:gate_asset_h3}
\end{figure*}

\begin{figure*}[t]
    \centering
    \includegraphics[width=\linewidth]{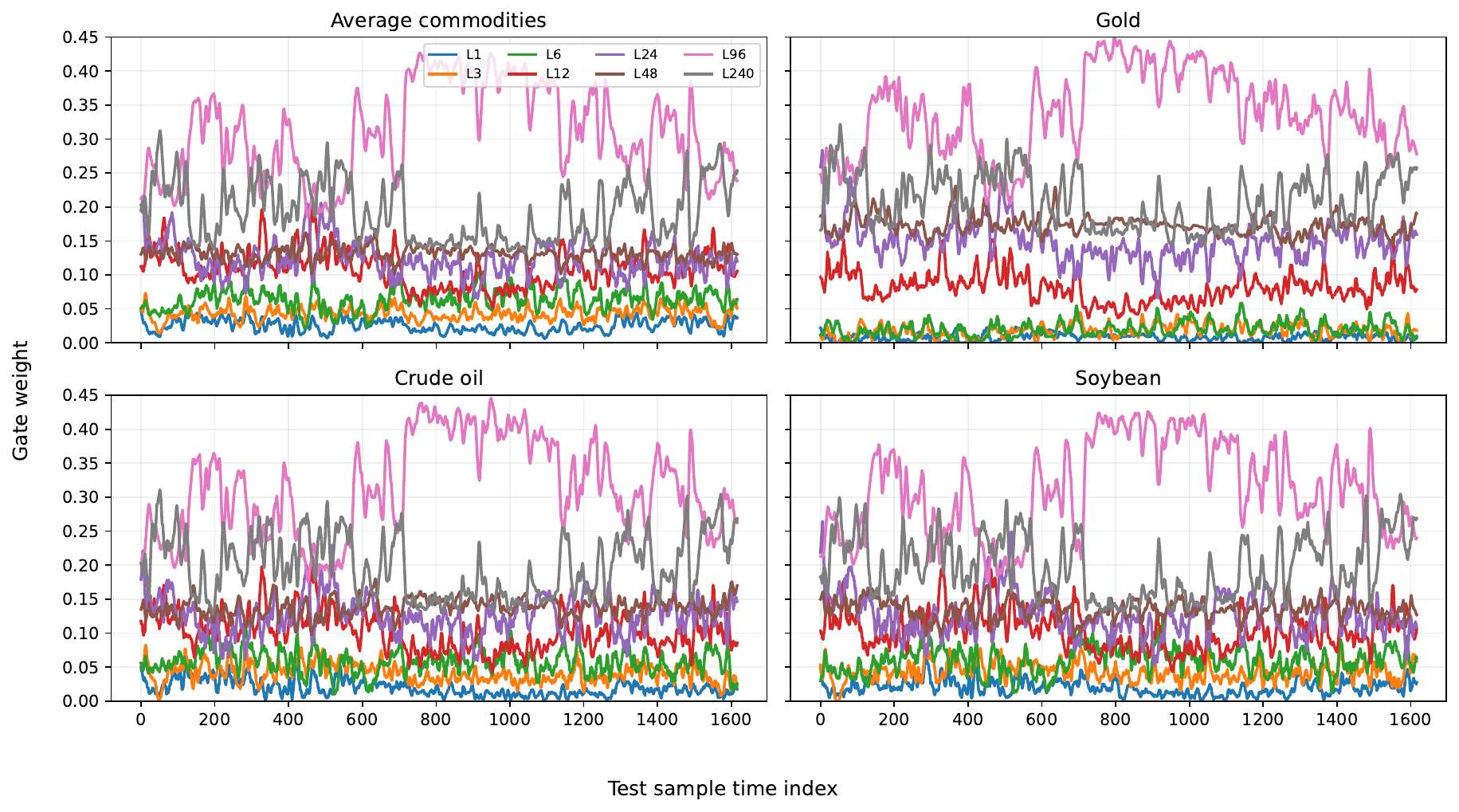}
    \caption{
    Learned asset-specific gate dynamics for forecasting horizon $h=12$.
    Commodity-specific temporal preferences remain distinct and evolve
    throughout the forecasting period.
    }
    \label{fig:gate_asset_h12}
\end{figure*}

\begin{figure*}[t]
    \centering
    \includegraphics[width=\linewidth]{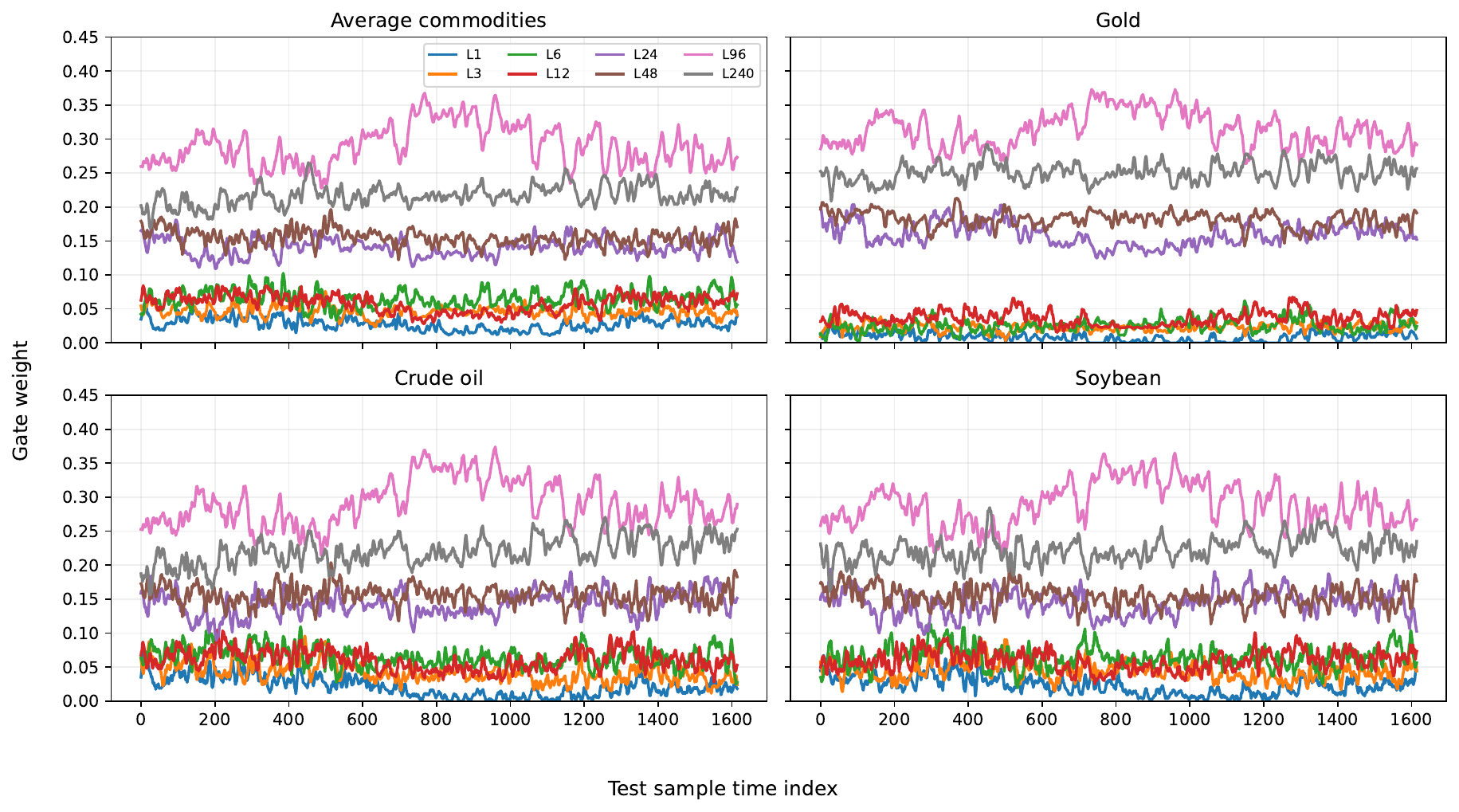}
    \caption{
    Learned asset-specific gate dynamics for forecasting horizon $h=24$.
    Different commodities continue to adaptively emphasize different temporal
    experts as market conditions evolve.
    }
    \label{fig:gate_asset_h24}
\end{figure*}

\begin{figure*}[t]
    \centering
    \includegraphics[width=\linewidth]{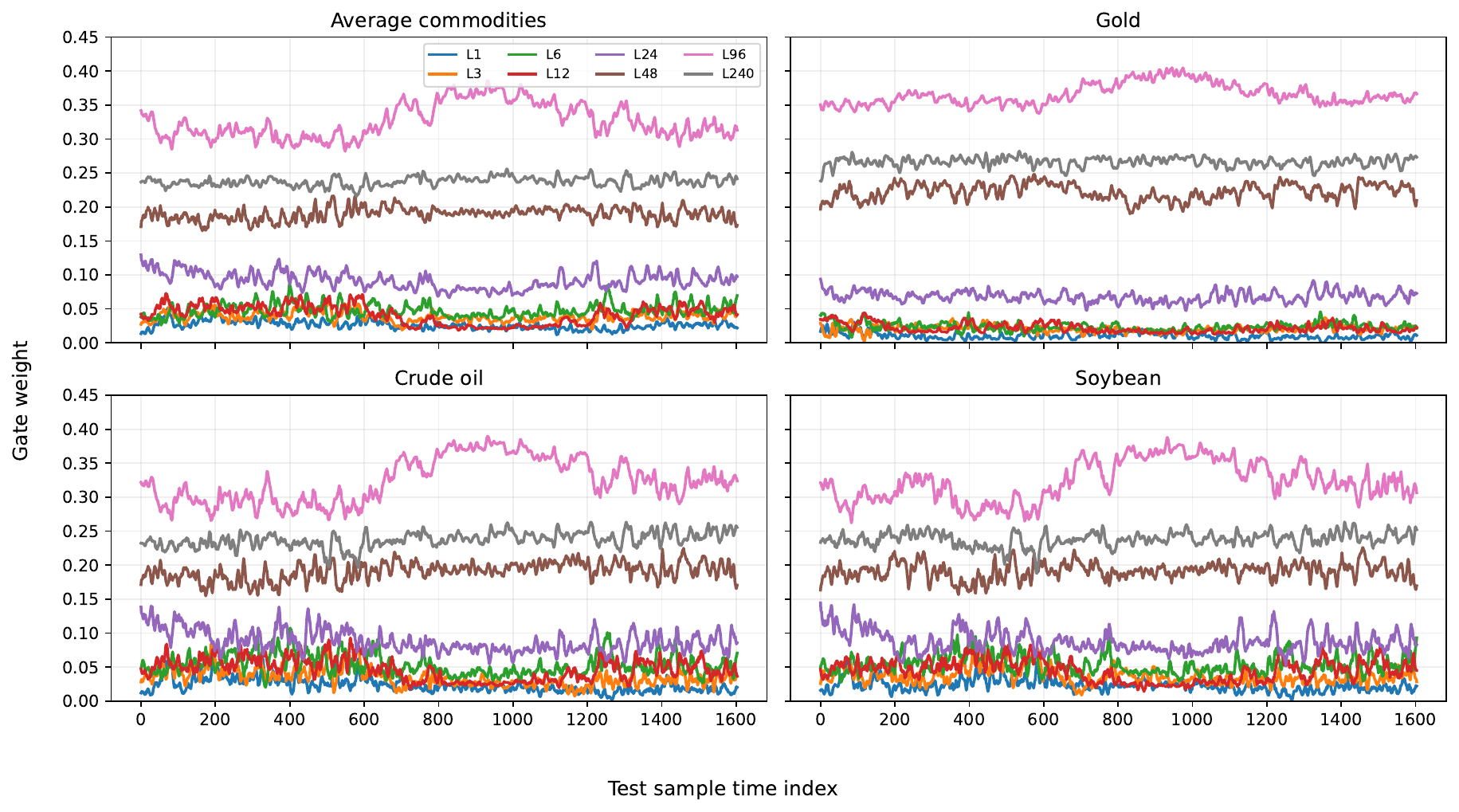}
    \caption{
    Learned asset-specific gate dynamics for forecasting horizon $h=96$.
    Even for the longest forecasting horizon, commodities continue to exhibit
    distinct and evolving temporal preferences.
    }
    \label{fig:gate_asset_h96}
\end{figure*}

Figures~\ref{fig:gate_asset_h3}, \ref{fig:gate_asset_h12}, \ref{fig:gate_asset_h24}, \ref{fig:gate_asset_h96} visualize the learned
gate dynamics for forecasting horizons
$h\in\{3,12,24,96\}$. For each horizon, the first panel reports the average
gate weights across all commodities, providing a market-level summary of the
learned forecasting regime. The remaining panels show three representative
commodities (gold, crude oil, and soybean), illustrating how temporal scale
preferences evolve for individual assets.

Across all forecasting horizons, two consistent patterns emerge. First, the
commodity-specific gates differ substantially from one another, indicating that
different commodities rely on different temporal scales even when observed at
the same point in time. Second, the preferred temporal scales vary continuously
throughout the test period, demonstrating that forecasting regimes evolve as
market conditions change. These observations suggest that both cross-commodity
and temporal predictive heterogeneity are intrinsic characteristics of the
forecasting problem.

\begin{figure*}[t]
    \centering
    \includegraphics[width=\linewidth]{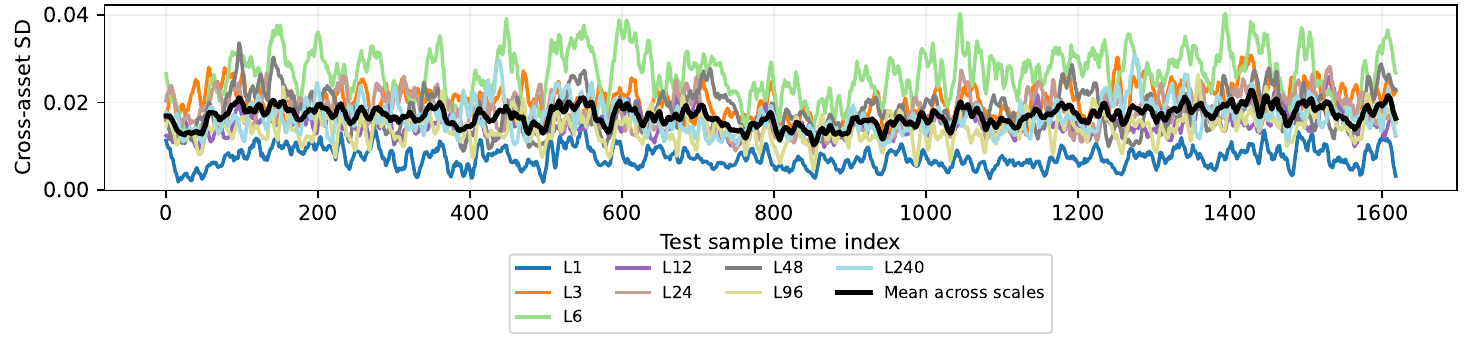}
    \caption{
    Cross-asset standard deviation of gate weights over time for forecasting
    horizon $h=3$. Larger values indicate stronger predictive heterogeneity
    across commodities.
    }
    \label{fig:gate_sd_h3}
\end{figure*}

\begin{figure*}[t]
    \centering
    \includegraphics[width=\linewidth]{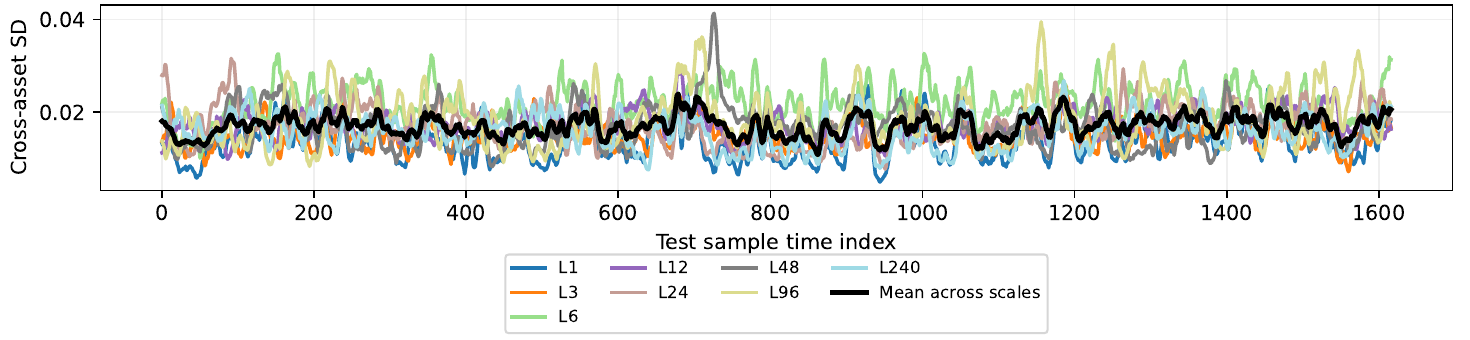}
    \caption{
    Cross-asset standard deviation of gate weights over time for forecasting
    horizon $h=12$, illustrating persistent predictive heterogeneity across
    commodities.
    }
    \label{fig:gate_sd_h12}
\end{figure*}

\begin{figure*}[t]
    \centering
    \includegraphics[width=\linewidth]{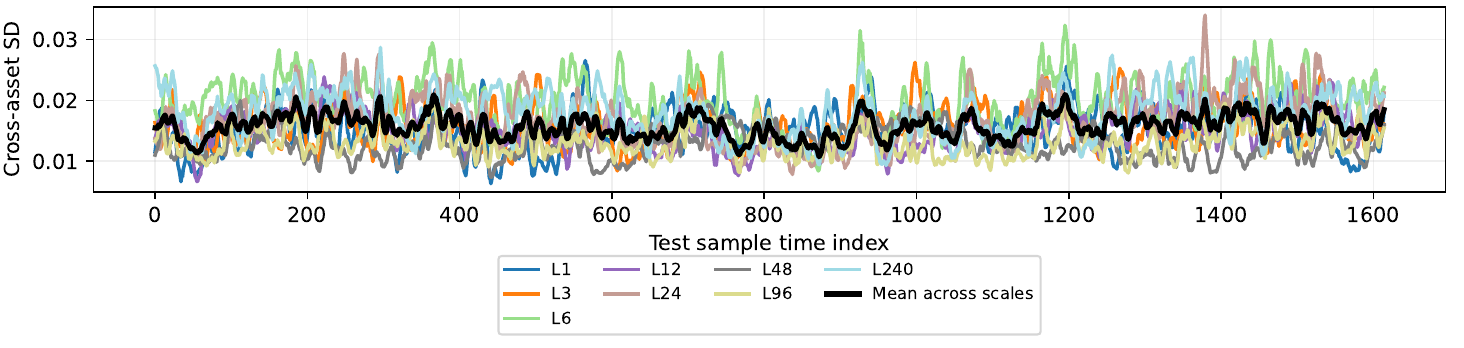}
    \caption{
    Cross-asset standard deviation of gate weights over time for forecasting
    horizon $h=24$, demonstrating substantial heterogeneity in temporal scale
    selection across commodities.
    }
    \label{fig:gate_sd_h24}
\end{figure*}

\begin{figure*}[t]
    \centering
    \includegraphics[width=\linewidth]{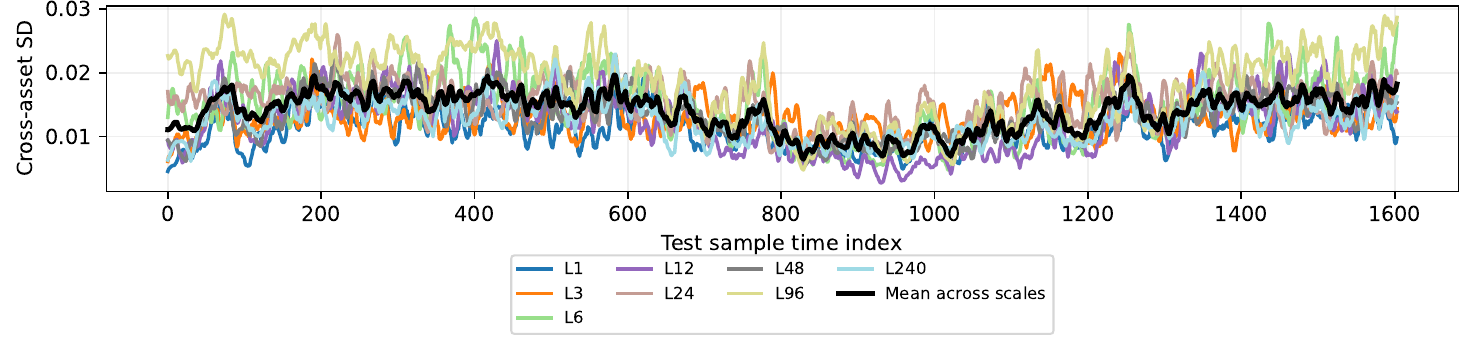}
    \caption{
    Cross-asset standard deviation of gate weights over time for forecasting
    horizon $h=96$, showing that predictive heterogeneity persists even for
    long-range forecasting.
    }
    \label{fig:gate_sd_h96}
\end{figure*}

The corresponding cross-asset heterogeneity is quantified in
Figures~\ref{fig:gate_sd_h3}, \ref{fig:gate_sd_h12}, \ref{fig:gate_sd_h24} \ref{fig:gate_sd_h96}, which report the standard
deviation of the learned gate weights across all commodities over time. Larger
dispersion indicates stronger disagreement among commodities regarding the most
informative temporal scale. Across all forecasting horizons, the dispersion
remains consistently positive and varies over time, confirming that predictive
heterogeneity is persistent rather than arising from random fluctuations.
Periods of elevated dispersion correspond to intervals in which commodities
adopt increasingly different temporal forecasting strategies, whereas lower
dispersion indicates more homogeneous market behavior.

Together, these empirical observations provide direct support for the design of
ABF. Because both the optimal forecasting scale and the degree of predictive
heterogeneity evolve across commodities and over time, a fixed single-scale
model cannot be expected to perform consistently well. At the same time,
indiscriminately averaging all temporal experts may unnecessarily combine strong
and weak predictors. The proposed asset-specific gating mechanism addresses this
trade-off by dynamically identifying the most informative temporal experts for
each commodity and forecasting time, while the validation-selected forecasting
anchor ensures that adaptive specialization improves forecasting performance
without introducing unnecessary estimation variance.

\subsection{Sensitivity analysis: localization bandwidth $\tau$}

The localization bandwidth $\tau$ controls the degree of similarity required
when selecting calibration residuals for GLCP. In the main experiments, $\tau$
is selected using the calibration data only, without accessing the test set.
After selecting $\tau$, the final GLCP intervals are constructed using the
complete calibration set and evaluated on the test set.

\begin{figure*}[t]
    \centering
    \includegraphics[width=0.7\linewidth]{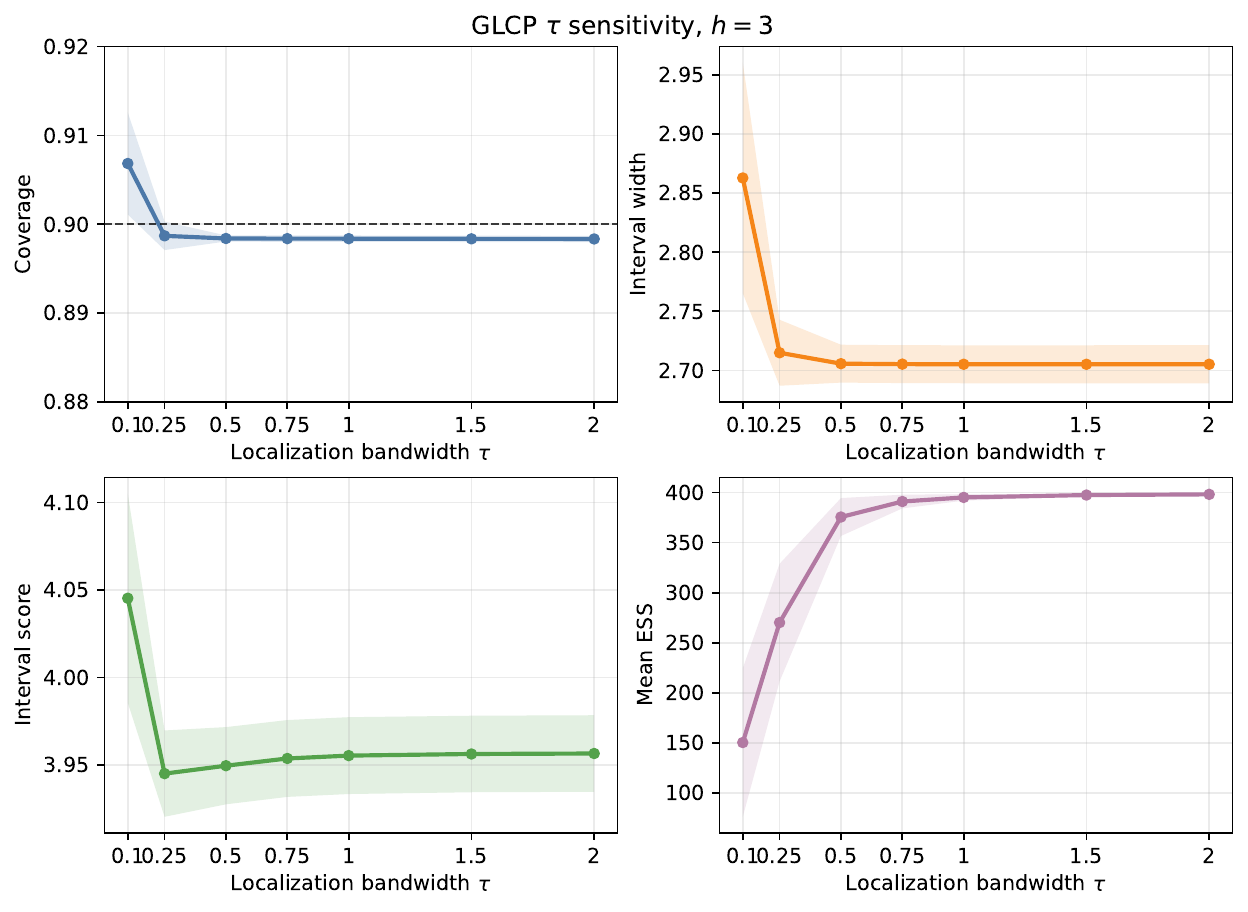}
    \caption{
    Sensitivity of GLCP to the localization bandwidth $\tau$ for forecasting
    horizon $h=3$. Each point corresponds to GLCP evaluated on the
    \textsc{Test} set using a fixed value of $\tau$.
    }
    \label{fig:tau_h3}
\end{figure*}

\begin{figure*}[t]
    \centering
    \includegraphics[width=0.7\linewidth]{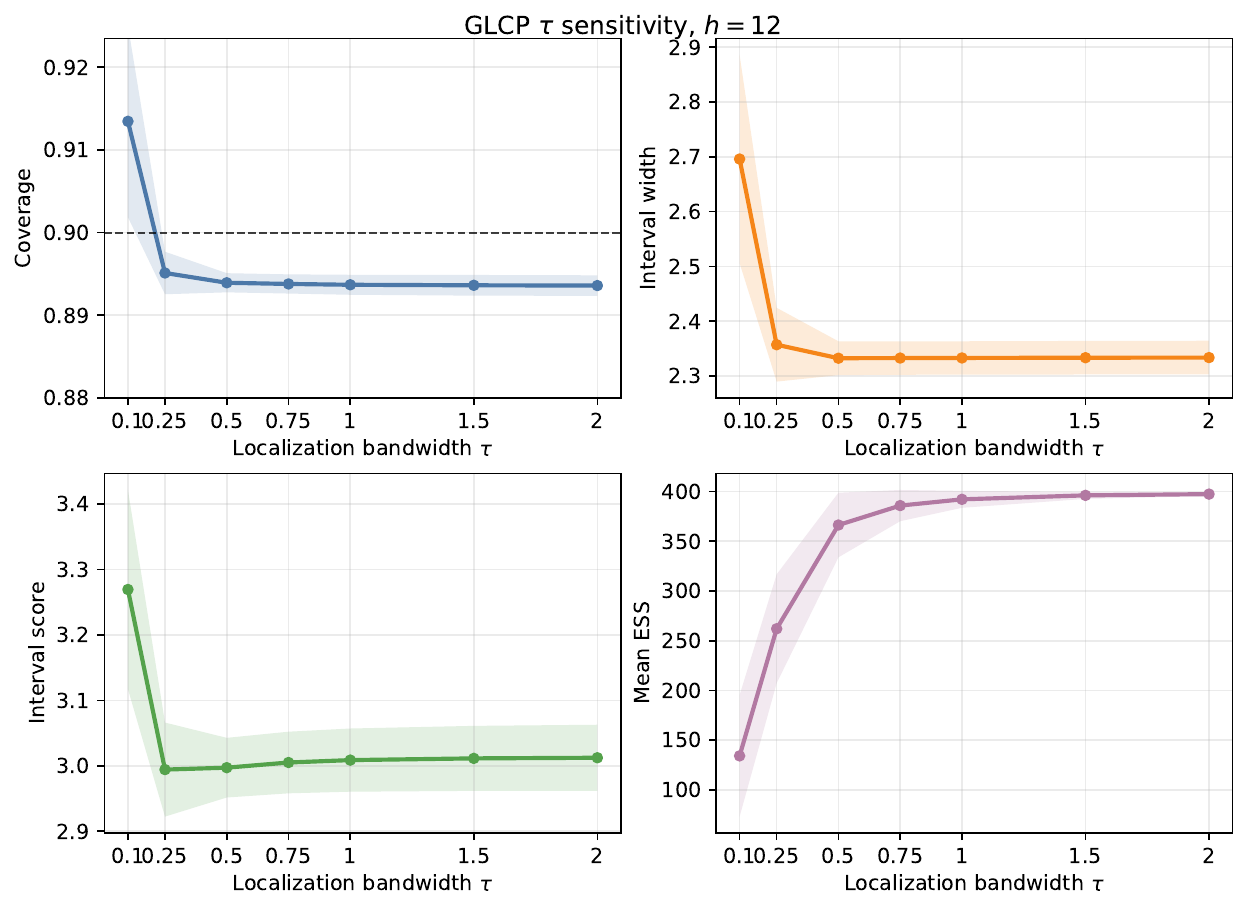}
    \caption{
    Sensitivity of GLCP to the localization bandwidth $\tau$ for forecasting
    horizon $h=12$.
    }
    \label{fig:tau_h12}
\end{figure*}

\begin{figure*}[t]
    \centering
    \includegraphics[width=0.7\linewidth]{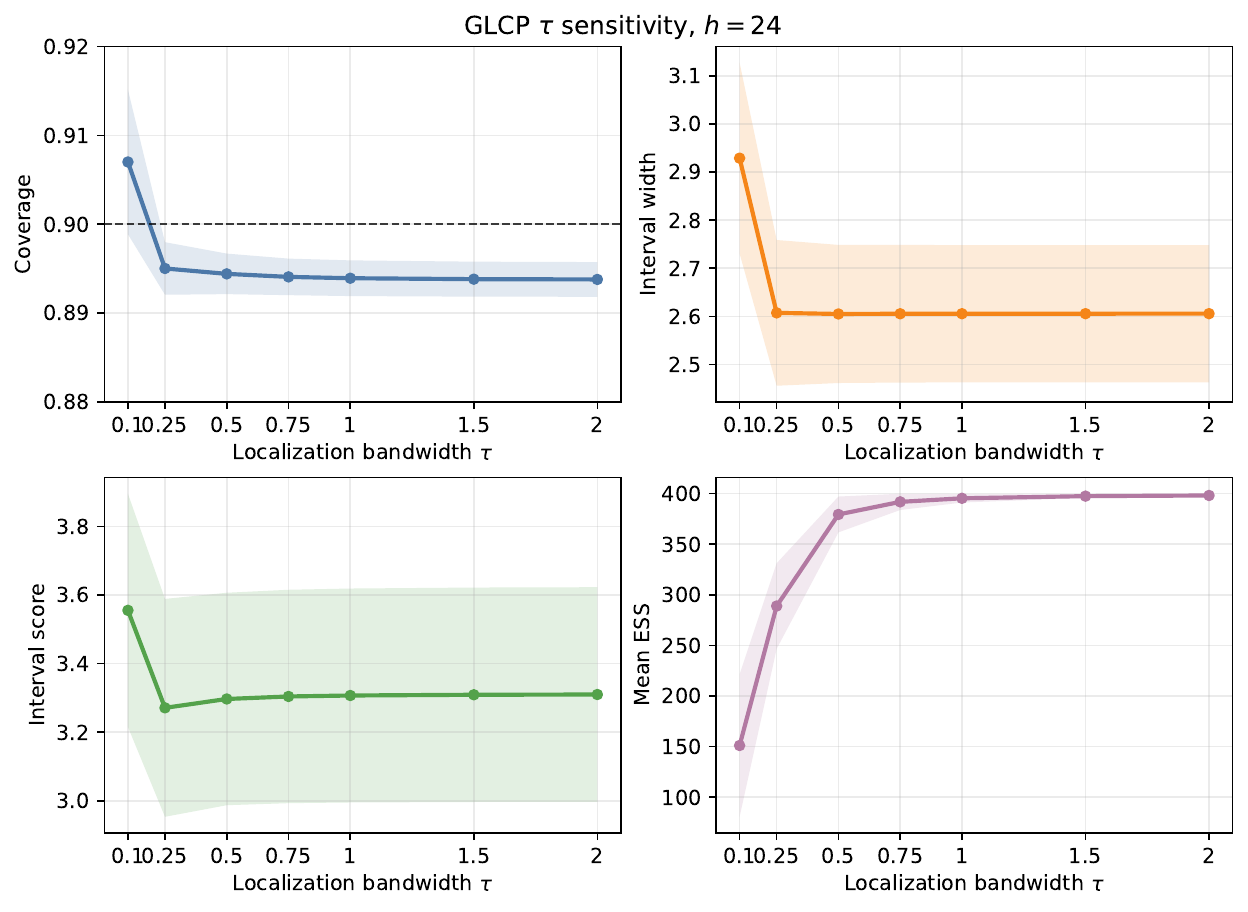}
    \caption{
    Sensitivity of GLCP to the localization bandwidth $\tau$ for forecasting
    horizon $h=24$.
    }
    \label{fig:tau_h24}
\end{figure*}

\begin{figure*}[t]
    \centering
    \includegraphics[width=0.7\linewidth]{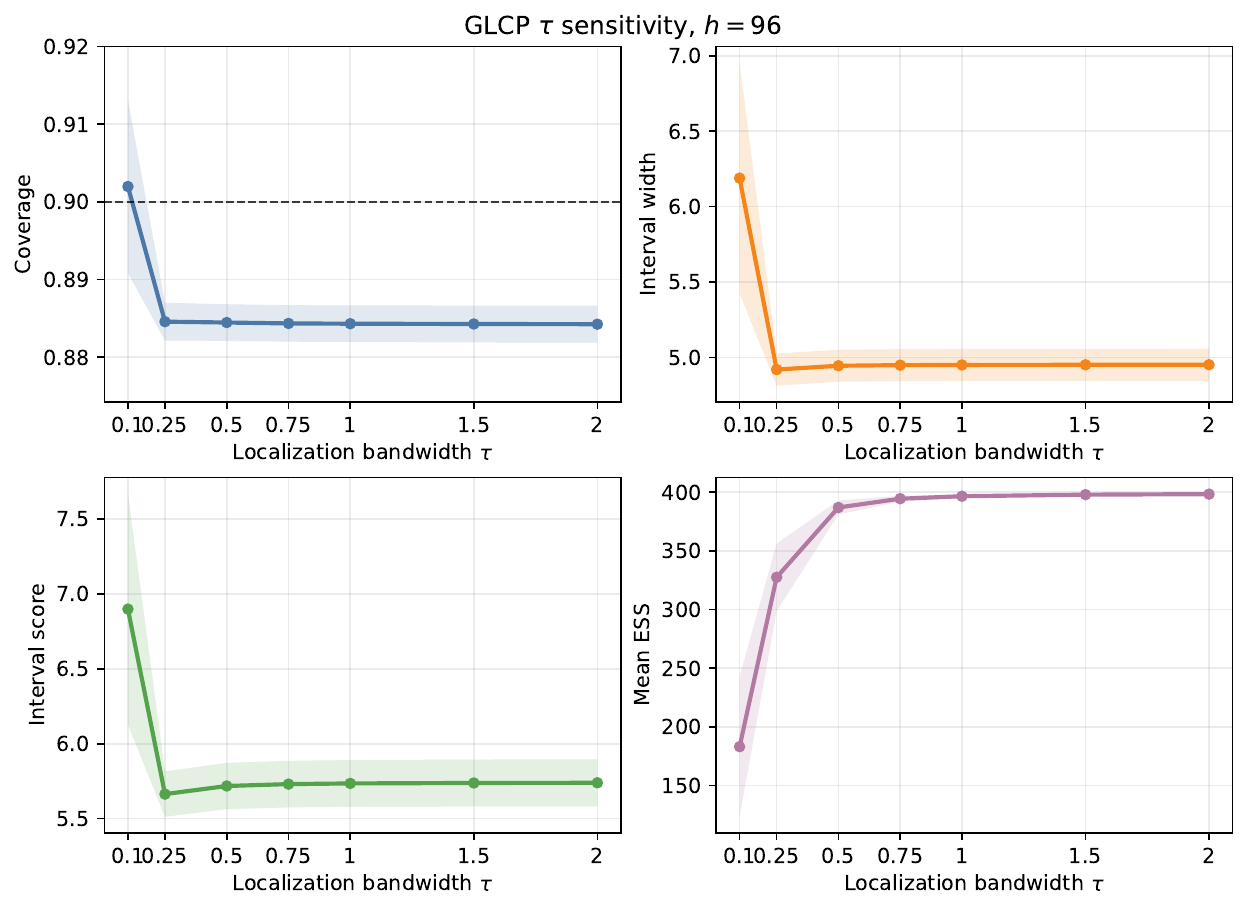}
    \caption{
    Sensitivity of GLCP to the localization bandwidth $\tau$ for forecasting
    horizon $h=96$.
    }
    \label{fig:tau_h96}
\end{figure*}

To further assess the robustness of GLCP with respect to this hyperparameter,
we additionally evaluate a range of fixed candidate values of $\tau$ on the
\textsc{Test} set. These experiments are reported solely as a sensitivity
analysis and are not used for model selection. The main paper presents the
results for the representative forecasting horizon $h=48$, while the remaining
forecasting horizons are provided here from Fig~\ref{fig:tau_h3}--\ref{fig:tau_h96}.

\subsection{Sensitivity analysis: time recency parameter $\lambda$}
We further examine the sensitivity of asymmetric GLCP to the temporal decay parameter $\lambda$, which controls the strength of recency weighting in the calibration set. Table~\ref{tab:lambda_sensitivity_h48_asym} reports results at the representative horizon $h=48$. The default value $\lambda=0.995$ achieves the lowest interval score while maintaining coverage close to the nominal level. A smaller value, $\lambda=0.990$, places stronger emphasis on very recent observations and yields substantially wider intervals, suggesting reduced effective calibration stability. In contrast, larger values such as $\lambda=0.9975$ and $\lambda=1$ use increasingly diffuse historical information; this increases the effective sample size but leads to less efficient intervals under nonstationary volatility dynamics. These results support the use of an intermediate decay rate, which balances adaptation to recent market conditions with sufficient calibration sample stability.
\begin{table*}[t]
\centering
 
\begin{tabular}{lcccc}
\toprule
$\lambda$ & Coverage & Width & IS & ESS \\
\midrule
0.99 & 0.925 (0.019) & 8.130 (3.811) & 8.502 (3.800) & 113.322 (59.101) \\
0.995 & 0.908 (0.002) & 3.064 (0.140) & \textbf{3.484} (0.153) & 327.257 (41.210) \\
0.9975 & 0.914 (0.002) & 3.735 (0.851) & 4.200 (0.861) & 517.660 (146.111) \\
1 & 0.932 (0.014) & 6.493 (2.158) & 6.979 (2.081) & 1287.315 (432.500) \\
\bottomrule
\end{tabular}
\caption{Sensitivity of asymmetric GLCP to the temporal decay parameter $\lambda$ at horizon $h=48$. Entries report mean (standard deviation) over 20 seeds; lower interval score and width are better, while coverage is targeted at $0.90$.}
\label{tab:lambda_sensitivity_h48_asym}
\end{table*}

\subsection{Learned sparse predictive transfer structure}

The sparse predictive transfer stage exploits residual cross-series information
after the dominant within-series temporal dynamics have been captured by ABF.
For each target series, the transfer model predicts the remaining ABF residual
using gate-weighted source-scale forecasts from related series. The learned
transfer coefficients therefore summarize which source series provide useful
residual information for each target series, after validation-based shrinkage
and negative-transfer control.

\begin{figure*}[t]
\centering
\includegraphics[width=0.85\linewidth]{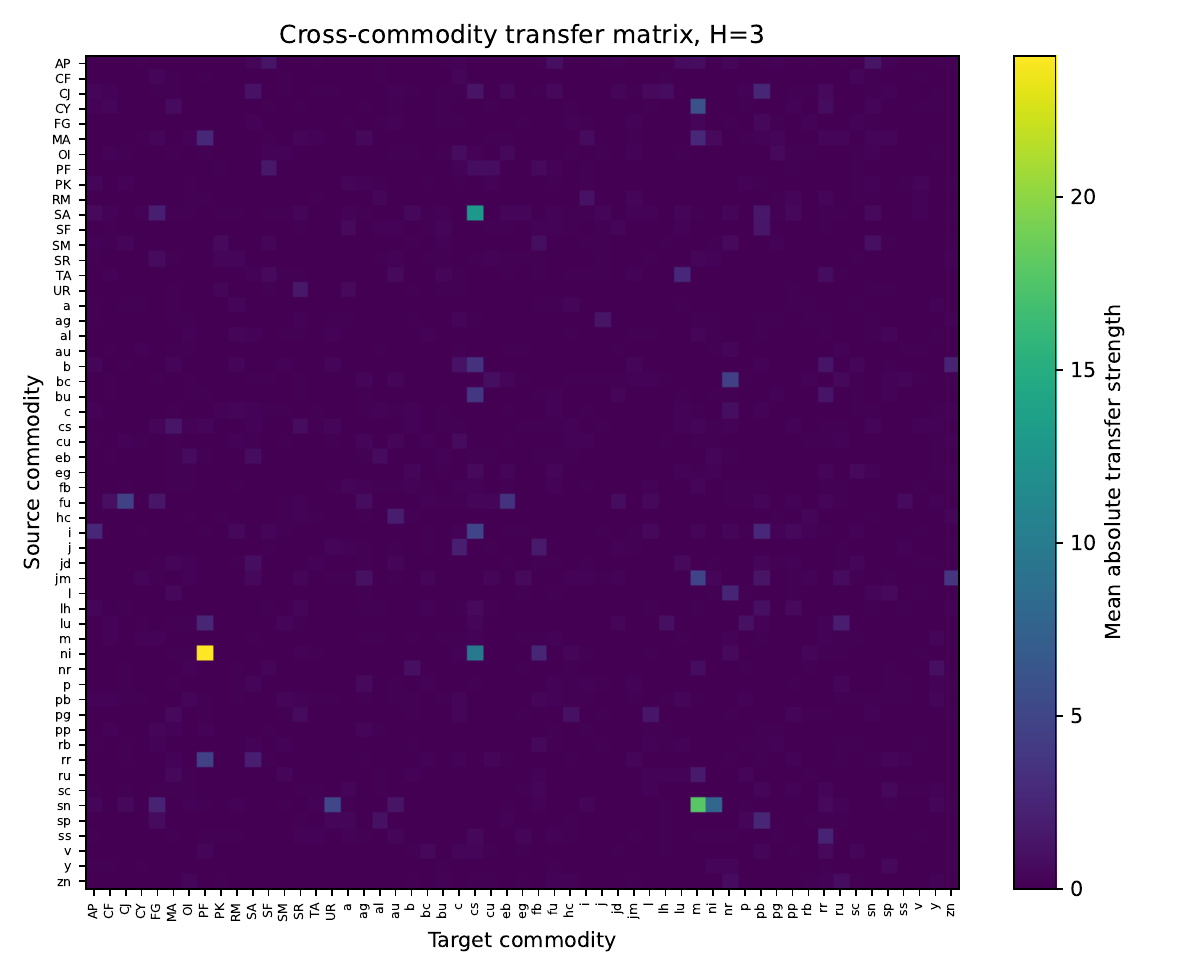}
\caption{
Learned sparse predictive transfer matrix for forecasting horizon $h=3$.
Rows correspond to target series and columns correspond to source series.
Each entry summarizes the learned transfer strength after aggregating over
gate-weighted scale-specific transfer features.
}
\label{fig:transfer_matrix_h3}
\end{figure*}

\begin{figure*}[t]
\centering
\includegraphics[width=0.85\linewidth]{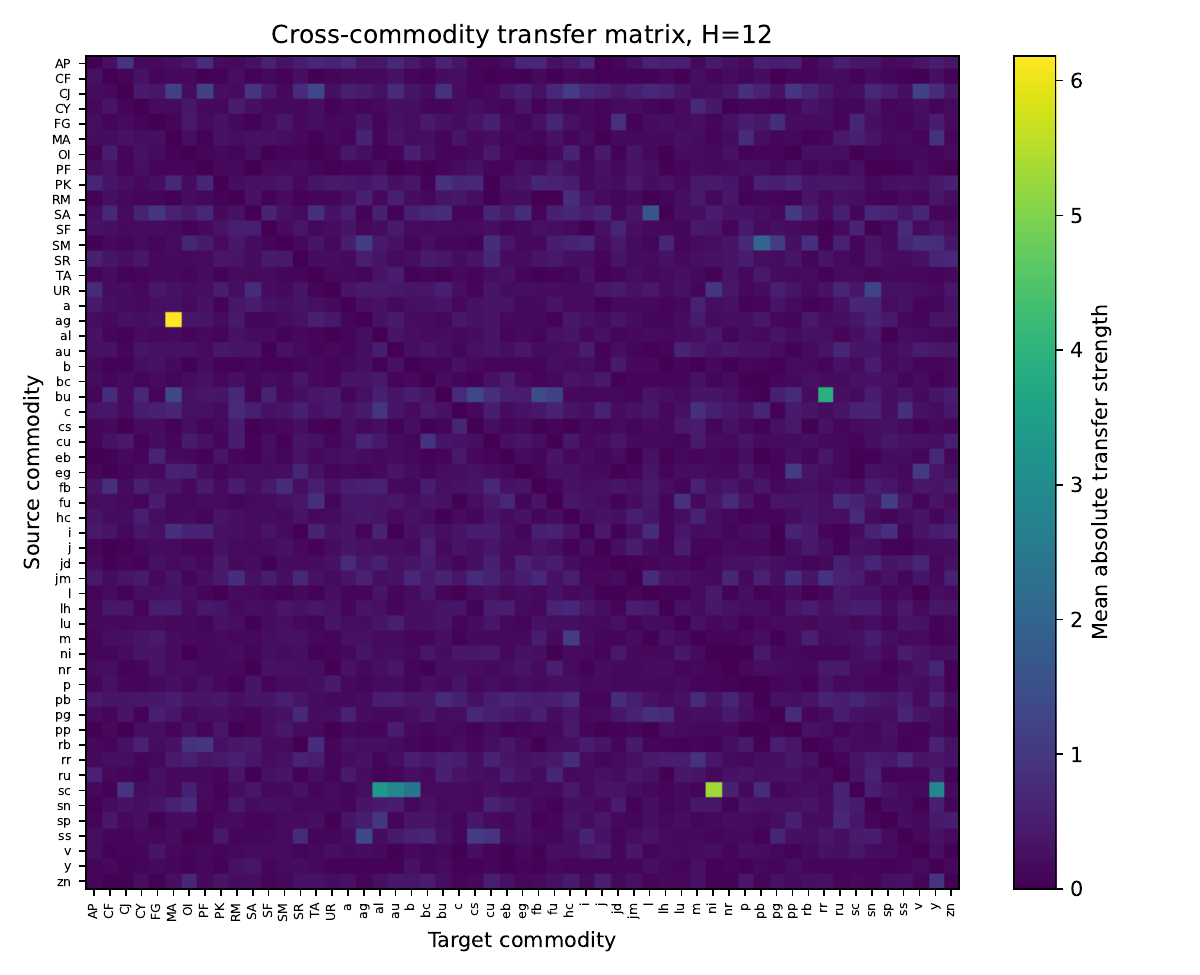}
\caption{
Learned sparse predictive transfer matrix for forecasting horizon $h=12$.
Rows correspond to target series and columns correspond to source series.
Each entry summarizes the learned transfer strength after aggregating over
gate-weighted scale-specific transfer features.
}
\label{fig:transfer_matrix_h12}
\end{figure*}

\begin{figure*}[t]
\centering
\includegraphics[width=0.85\linewidth]{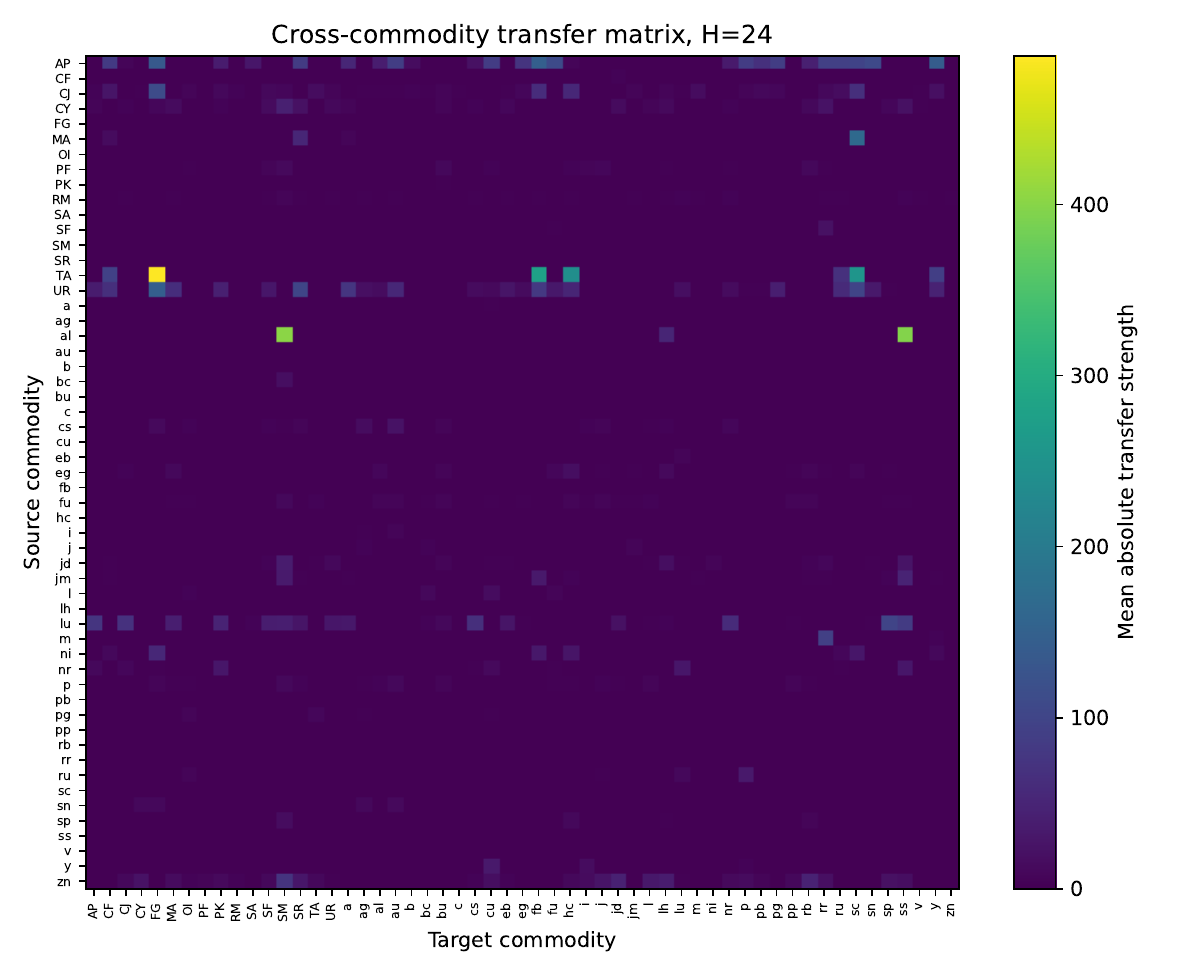}
\caption{
Learned sparse predictive transfer matrix for forecasting horizon $h=24$.
Rows correspond to target series and columns correspond to source series.
Each entry summarizes the learned transfer strength after aggregating over
gate-weighted scale-specific transfer features.
}
\label{fig:transfer_matrix_h24}
\end{figure*}

\begin{figure*}[t]
\centering
\includegraphics[width=0.85\linewidth]{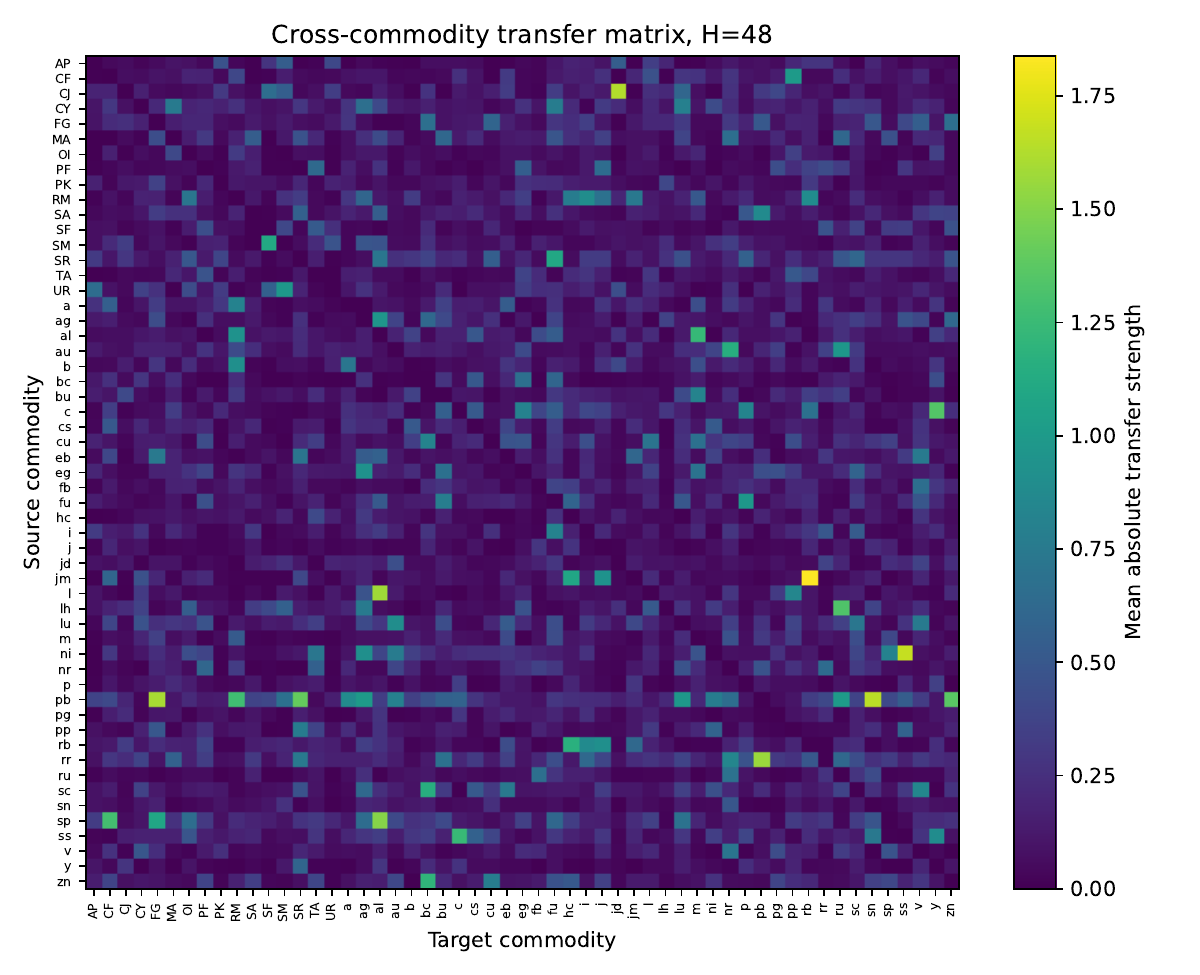}
\caption{
Learned sparse predictive transfer matrix for forecasting horizon $h=48$.
Rows correspond to target series and columns correspond to source series.
Each entry summarizes the learned transfer strength after aggregating over
gate-weighted scale-specific transfer features.
}
\label{fig:transfer_matrix_h48}
\end{figure*}

\begin{figure*}[t]
\centering
\includegraphics[width=0.85\linewidth]{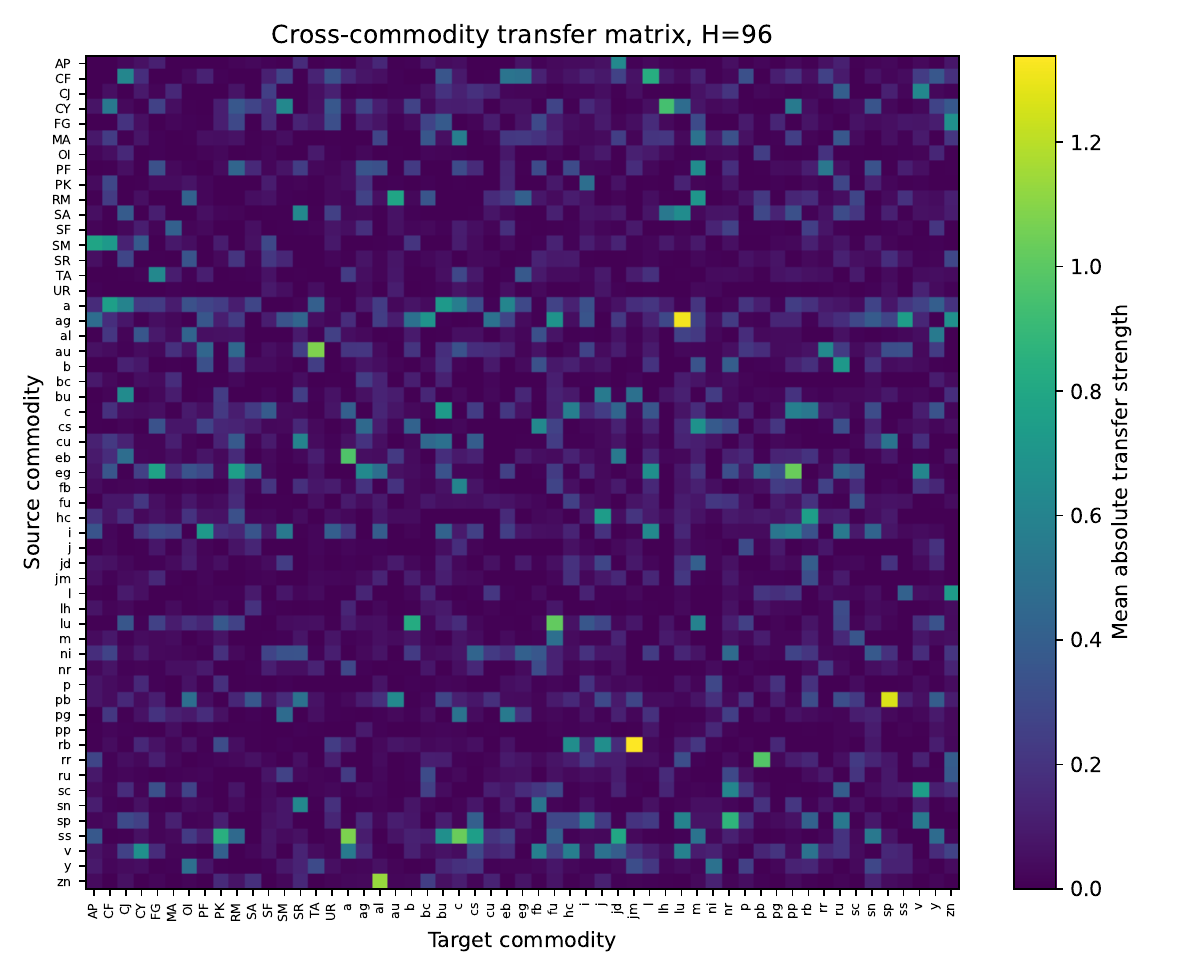}
\caption{
Learned sparse predictive transfer matrix for forecasting horizon $h=96$.
Rows correspond to target series and columns correspond to source series.
Each entry summarizes the learned transfer strength after aggregating over
gate-weighted scale-specific transfer features.
}
\label{fig:transfer_matrix_h96}
\end{figure*}

Figures~\ref{fig:transfer_matrix_h3}--\ref{fig:transfer_matrix_h96}
visualize the learned transfer matrices across forecasting horizons. Each
matrix summarizes the strength of the sparse transfer correction from source
series to target series, aggregated over the scale-specific transfer features.
The sparse and horizon-dependent patterns indicate that useful cross-series
residual structure is selective rather than uniform across all commodities.
This supports the use of validation-controlled sparse transfer instead of
indiscriminate cross-series pooling.

\end{document}